%% file: main.tex
\pdfoutput=1
\documentclass{article}

\usepackage{PRIMEarxiv}

\usepackage[utf8]{inputenc}
\usepackage[T1]{fontenc}
\usepackage{amsmath}
\usepackage{booktabs}
\usepackage{array}
\usepackage{tabularx}
\usepackage{microtype}
\usepackage{graphicx}
\usepackage{xspace}
\xspaceaddexceptions{\%}
\usepackage[round]{natbib}
\usepackage{xurl}
\usepackage{hyperref}
\hypersetup{colorlinks=true, linkcolor=blue, citecolor=blue, urlcolor=blue}
\usepackage[capitalise,nameinlink]{cleveref}
\hypersetup{pdftitle={Worth the Wait? How Response Time and Streaming Shape Judgments of Generative AI Answers and Chatbot Engagement},
  pdfauthor={Zihan Chen, Di Zhu, Lei Nico Zheng},
  pdfkeywords={response time, latency, generative AI, large language models, chatbots, streaming, caching, labor illusion, operational transparency, user engagement, AI evaluation, human-AI interaction}}
\graphicspath{{figures/}}

\input{generated/numbers}

\title{Worth the Wait?\\How Response Time and Streaming Shape Judgments of Generative AI Answers and Chatbot Engagement}

\author{
  Zihan Chen \\
  Stevens Institute of Technology \\
  Hoboken, NJ, USA \\
  \texttt{zchen61@stevens.edu} \\
  \And
  Di Zhu \\
  Stevens Institute of Technology \\
  Hoboken, NJ, USA \\
  \texttt{dzhu1@stevens.edu} \\
  \And
  Lei Nico Zheng \\
  University of Massachusetts Boston \\
  Boston, MA, USA \\
  \texttt{lei.zheng@umb.edu} \\
}

\begin{document}
\maketitle

\input{sections/00_abstract}
\input{sections/01_intro}
\input{sections/02_background}
\input{sections/03_setting}
\input{sections/04_study1}
\input{sections/05_study2}
\input{sections/06_study3}
\input{sections/07_discussion}

\bibliographystyle{plainnat}
\bibliography{references}

\appendix
\input{sections/08_appendix}

\end{document}

%% file: generated/numbers.tex
\newcommand{\SoneEquivLchAbs}{1.3\xspace}
\newcommand{\SoneEquivLchSE}{0.7\xspace}
\newcommand{\SoneEquivLtokAbs}{2.8\xspace}
\newcommand{\SoneEquivLtokSE}{1.0\xspace}
\newcommand{\SoneEquivLreasAbs}{4.7\xspace}
\newcommand{\SoneEquivLreasSE}{1.7\xspace}
\newcommand{\SoneTieAbsdcAbs}{9.8\xspace}
\newcommand{\SoneTieAbsdcSE}{2.6\xspace}
\newcommand{\SoneTieAbsdcMean}{36.1\xspace}
\newcommand{\SoneVoteVOneAbs}{26.2\xspace}
\newcommand{\SoneVoteVTwoAbs}{23.7\xspace}
\newcommand{\SoneVoteVThreeB}{$-$25.8\xspace}
\newcommand{\SoneVoteVThreeAbs}{25.8\xspace}
\newcommand{\SoneVoteVThreeSE}{4.4\xspace}
\newcommand{\SoneVoteVFourAbs}{25.7\xspace}
\newcommand{\SoneTagCompleteAbs}{21.0\xspace}
\newcommand{\SoneTagCompleteSE}{5.2\xspace}
\newcommand{\SoneTagIncorrectB}{+8.0\xspace}
\newcommand{\SoneTagIncorrectAbs}{8.0\xspace}
\newcommand{\SoneTagIncorrectSE}{2.9\xspace}
\newcommand{\SoneTagSuperficialAbs}{12.7\xspace}
\newcommand{\SoneTagSuperficialSE}{4.0\xspace}
\newcommand{\SoneRobNotopAbs}{24.3\xspace}
\newcommand{\SoneRobTimesensB}{$-$24.7\xspace}
\newcommand{\SoneRobTimesensAbs}{24.7\xspace}
\newcommand{\SoneRobPopularAbs}{24.3\xspace}
\newcommand{\SoneRobNopooledAbs}{26.0\xspace}
\newcommand{\SoneRobVerifiedAbs}{27.9\xspace}
\newcommand{\SoneRobRegimeOneAbs}{24.1\xspace}
\newcommand{\SoneRobRegimeTwoAbs}{27.1\xspace}
\newcommand{\SoneHetShortAbs}{19.1\xspace}
\newcommand{\SoneHetMediumAbs}{33.8\xspace}
\newcommand{\SoneHetLongAbs}{48.1\xspace}
\newcommand{\SoneHetReasoningAbs}{19.4\xspace}
\newcommand{\SoneHetNoreasoningAbs}{28.8\xspace}
\newcommand{\SoneRxLikeshareAbs}{41.4\xspace}
\newcommand{\SoneRxLikeshareMpAbs}{32.2\xspace}
\newcommand{\SoneRxCellLikedCOneZeroAbs}{3.0\xspace}
\newcommand{\SoneRxCellLikedCOneZeroSE}{0.5\xspace}
\newcommand{\SoneRxCellLikedCZeroOneB}{$-$1.0\xspace}
\newcommand{\SoneRxCellLikedCZeroOneSE}{0.6\xspace}
\newcommand{\SoneRxCellLikedCOneOneB}{$-$1.4\xspace}
\newcommand{\SoneRxCellLikedCOneOneAbs}{1.4\xspace}
\newcommand{\SoneRxCellLikedCOneOneSE}{0.5\xspace}
\newcommand{\SoneRxCellDislikedCOneZeroAbs}{2.1\xspace}
\newcommand{\SoneRxCellDislikedCOneZeroSE}{0.5\xspace}
\newcommand{\SoneRxCellDislikedCZeroOneB}{+0.3\xspace}
\newcommand{\SoneRxCellDislikedCZeroOneSE}{0.3\xspace}
\newcommand{\SoneRxCellDislikedCOneOneAbs}{0.9\xspace}
\newcommand{\SoneRxCellDislikedCOneOneSE}{0.4\xspace}
\newcommand{\SoneEngContB}{+0.8\xspace}
\newcommand{\SoneEngContSE}{0.6\xspace}
\newcommand{\SoneEngContMean}{8.8\xspace}
\newcommand{\SoneIdentDecisiveAbs}{29.8\xspace}
\newcommand{\SoneIdentDecisiveSE}{2.8\xspace}
\newcommand{\SoneIdentDecisiveEligAbs}{28.2\xspace}
\newcommand{\SoneVoteBotheligPhAbs}{30.3\xspace}
\newcommand{\SoneVoteBotheligPhSE}{5.6\xspace}
\newcommand{\SoneIdentEqwAbs}{27.3\xspace}
\newcommand{\SoneIdentEqwSE}{3.7\xspace}
\newcommand{\SoneIdentEqwN}{265\xspace}
\newcommand{\SoneIdentDecisiveCltextSE}{3.5\xspace}
\newcommand{\SoneIdentDecisiveEligCltextSE}{9.1\xspace}
\newcommand{\SoneIdentDecisiveOppAbs}{30.2\xspace}
\newcommand{\SoneIdentDecisiveFourEighthAbs}{28.0\xspace}
\newcommand{\SoneIdentTagCompleteAbs}{23.2\xspace}
\newcommand{\SoneIdentTagCompleteSE}{3.5\xspace}
\newcommand{\SoneIdentTagSuperficialAbs}{12.2\xspace}
\newcommand{\SoneIdentTagSuperficialSE}{3.0\xspace}
\newcommand{\SoneIdentTagIncorrectAbs}{7.0\xspace}
\newcommand{\SoneIdentTagIncorrectSE}{2.1\xspace}
\newcommand{\SoneRobHitAbs}{25.4\xspace}
\newcommand{\SoneRobHitSE}{10.1\xspace}
\newcommand{\SoneRobWTwoFourAbs}{24.0\xspace}
\newcommand{\SoneRobWTwoFourSE}{4.6\xspace}
\newcommand{\StwoContOlsAbs}{1.8\xspace}
\newcommand{\StwoContOlsSE}{0.2\xspace}
\newcommand{\StwoContOlsN}{324{,}029\xspace}
\newcommand{\StwoContIvAbs}{2.0\xspace}
\newcommand{\StwoContIvSE}{1.0\xspace}
\newcommand{\StwoContIvN}{324{,}029\xspace}
\newcommand{\StwoContIvF}{1{,}066\xspace}
\newcommand{\StwoVotedIvB}{0.0\xspace}
\newcommand{\StwoVotedIvSE}{0.8\xspace}
\newcommand{\StwoReactedIvB}{+1.7\xspace}
\newcommand{\StwoReactedIvSE}{0.7\xspace}
\newcommand{\StwoContOlsVisAbs}{1.5\xspace}
\newcommand{\StwoContOlsVisSE}{0.2\xspace}
\newcommand{\StwoContOlsVisN}{280{,}993\xspace}
\newcommand{\StwoContIvVisAbs}{2.4\xspace}
\newcommand{\StwoContIvVisSE}{1.1\xspace}
\newcommand{\StwoContIvVisN}{280{,}993\xspace}
\newcommand{\StwoContIvVisF}{973\xspace}
\newcommand{\StwoContkOlsAbs}{3.7\xspace}
\newcommand{\StwoContkOlsSE}{0.3\xspace}
\newcommand{\StwoContkOlsN}{140{,}671\xspace}
\newcommand{\StwoContkOlsVisAbs}{4.2\xspace}
\newcommand{\StwoContkOlsVisSE}{0.4\xspace}
\newcommand{\StwoContkOlsVisN}{121{,}794\xspace}
\newcommand{\StwoContkIvB}{$-$1.6\xspace}
\newcommand{\StwoContkIvSE}{1.5\xspace}
\newcommand{\StwoContkIvN}{140{,}368\xspace}
\newcommand{\StwoVoteOlsB}{$-$1.0\xspace}
\newcommand{\StwoVoteOlsSE}{0.4\xspace}
\newcommand{\StwoVoteOlsDlcB}{+13.9\xspace}
\newcommand{\StwoVoteOlsDlcAbs}{13.9\xspace}
\newcommand{\StwoVoteOlsDlcSE}{0.4\xspace}
\newcommand{\StwoVoteIvB}{$-$6.6\xspace}
\newcommand{\StwoVoteIvSE}{3.2\xspace}
\newcommand{\StwoRxpLikedOwnMean}{6.0\xspace}
\newcommand{\StwoRxpDislikedOwnMean}{2.8\xspace}
\newcommand{\StwoContRfB}{$-$1.8\xspace}
\newcommand{\StwoContRfSE}{0.9\xspace}
\newcommand{\StwoVoteIvOrderedB}{$-$2.9\xspace}
\newcommand{\StwoVoteIvOrderedSE}{2.5\xspace}
\newcommand{\StwoVoteOlsOrderedB}{+0.2\xspace}
\newcommand{\StwoVoteOlsOrderedSE}{0.4\xspace}
\newcommand{\StwoReturnSevendIvCensB}{$-$0.6\xspace}
\newcommand{\StwoReturnSevendIvCensSE}{0.5\xspace}
\newcommand{\StwoVoteOlsOpmB}{+0.2\xspace}
\newcommand{\StwoVoteIvOpmB}{$-$2.7\xspace}
\newcommand{\StwoContIvClwkSE}{1.1\xspace}
\newcommand{\StwoContOlsClwkSE}{0.2\xspace}
\newcommand{\StwoContIvRawexpB}{$-$2.0\xspace}
\newcommand{\StwoContIvRawexpSE}{1.0\xspace}
\newcommand{\SoneNConvEra}{57{,}859\xspace}
\newcommand{\SoneNConvElig}{6{,}898\xspace}
\newcommand{\SoneNConvBothElig}{4{,}801\xspace}
\newcommand{\SoneNEligAnswers}{11{,}699\xspace}
\newcommand{\SoneNEligCells}{2{,}460\xspace}
\newcommand{\SoneNEligPrompts}{1{,}077\xspace}
\newcommand{\SoneNVotesEligAll}{1{,}574\xspace}
\newcommand{\SoneNVotesDecisive}{1{,}005\xspace}
\newcommand{\SoneNVotesOnecached}{474\xspace}
\newcommand{\SoneShareCachedElig}{65\xspace}
\newcommand{\SoneShareCachedRegimeOne}{42\xspace}
\newcommand{\SoneShareCachedRegimeTwo}{74\xspace}
\newcommand{\SoneReplayMatchShare}{93\xspace}
\newcommand{\SoneRawWinCached}{22\xspace}
\newcommand{\SoneDurCachedMedian}{0.3\xspace}
\newcommand{\SoneDurCachedPSevenFive}{1.1\xspace}
\newcommand{\SoneDurFreshMedian}{4.3\xspace}
\newcommand{\SoneDurFreshPTwoFive}{1.4\xspace}
\newcommand{\SoneDurFreshPSevenFive}{20.3\xspace}
\newcommand{\SoneRxintLikedXIntB}{+2.56\xspace}
\newcommand{\SoneRxintLikedXIntSE}{0.62\xspace}
\newcommand{\SoneStart}{16 March 2026\xspace}
\newcommand{\SoneEnd}{18 May 2026\xspace}
\newcommand{\SoneRxLikedRel}{62\xspace}
\newcommand{\SoneRxDislikedRel}{86\xspace}
\newcommand{\SoneReplayLdurAbsLog}{1.7\xspace}
\newcommand{\SoneEngContCILo}{$-$0.3\xspace}
\newcommand{\SoneEngContCIHi}{+1.9\xspace}
\newcommand{\SoneAuditNEligNorm}{12{,}518\xspace}
\newcommand{\SoneAuditNEligExact}{12{,}057\xspace}
\newcommand{\SoneAuditNEligNormOnly}{461\xspace}
\newcommand{\SoneAuditNCached}{8{,}604\xspace}
\newcommand{\SoneAuditCachedIneligNorm}{509\xspace}
\newcommand{\SoneAuditIneligShare}{6\xspace}
\newcommand{\SoneCellsMixed}{474\xspace}
\newcommand{\SoneAnswersInMixedCells}{9{,}339\xspace}
\newcommand{\SoneNEligVisitors}{2{,}026\xspace}
\newcommand{\SoneIdentNTexts}{1{,}948\xspace}
\newcommand{\SoneIdentDecisiveEligTexts}{44\xspace}
\newcommand{\SoneIdentDecisiveBothTexts}{265\xspace}
\newcommand{\SoneTopPromptShare}{19\xspace}
\newcommand{\SoneTopOneZeroPromptShare}{62\xspace}
\newcommand{\SoneTopVotePromptShare}{6\xspace}
\newcommand{\SoneLeeDecLower}{$-$39.6\xspace}
\newcommand{\SoneLeeDecUpper}{$-$28.9\xspace}
\newcommand{\SoneLeeDecCiLo}{$-$52.4\xspace}
\newcommand{\SoneLeeDecCiHi}{$-$20.3\xspace}
\newcommand{\SoneLeeAnyLower}{$-$28.6\xspace}
\newcommand{\SoneLeeAnyUpper}{$-$17.0\xspace}
\newcommand{\SoneLeeAnyCiLo}{$-$36.0\xspace}
\newcommand{\SoneLeeAnyCiHi}{$-$8.9\xspace}
\newcommand{\SoneJointOWinAbs}{1.5\xspace}
\newcommand{\SoneJointOWinBase}{11.2\xspace}
\newcommand{\SoneJointOLoseAbs}{2.3\xspace}
\newcommand{\SoneJointOLoseBase}{11.0\xspace}
\newcommand{\SoneJointOTieB}{$-$0.5\xspace}
\newcommand{\SoneJointOTieSE}{0.4\xspace}
\newcommand{\SoneJointONovoteB}{$-$0.3\xspace}
\newcommand{\SoneJointONovoteSE}{0.6\xspace}
\newcommand{\SoneJointONovoteBase}{62.5\xspace}
\newcommand{\SoneJointOScoreAbs}{15.9\xspace}
\newcommand{\SoneJointOScoreSE}{2.3\xspace}
\newcommand{\SoneStateFFWinA}{50\xspace}
\newcommand{\SoneStateRRWinA}{49\xspace}
\newcommand{\SoneStateDiscordantReplayWin}{26\xspace}
\newcommand{\SoneSpillPct}{14\xspace}
\newcommand{\SoneIdentdNBothFresh}{310\xspace}
\newcommand{\SoneIdentdNBothReplay}{334\xspace}
\newcommand{\SoneIdentdNBothVis}{445\xspace}
\newcommand{\SoneIdentdNBothPrompts}{118\xspace}
\newcommand{\SoneIdentdHSinceMedian}{0.9\xspace}
\newcommand{\SoneIdentEqwFresh}{57\xspace}
\newcommand{\SoneIdentEqwReplay}{29\xspace}
\newcommand{\SoneIdentReasAnyShare}{28\xspace}
\newcommand{\SoneIdentReasSameShare}{88\xspace}
\newcommand{\SoneLenPriorB}{$-$3.1\xspace}
\newcommand{\SoneLenPriorSE}{2.8\xspace}
\newcommand{\SoneLenPriorNophB}{$-$2.5\xspace}
\newcommand{\SoneLenPriorNophSE}{2.2\xspace}
\newcommand{\SoneRxCellbaseLiked}{5.6\xspace}
\newcommand{\SoneRxCellbaseDisliked}{1.5\xspace}
\newcommand{\SoneLeeDecROneLower}{$-$41.8\xspace}
\newcommand{\SoneLeeDecROneUpper}{$-$22.9\xspace}
\newcommand{\SoneLeeDecRTwoLower}{$-$37.8\xspace}
\newcommand{\SoneLeeDecRTwoUpper}{$-$33.2\xspace}
\newcommand{\SoneSpillImpliedLo}{$-$1.1\xspace}
\newcommand{\SoneSpillImpliedHi}{+0.3\xspace}
\newcommand{\StwoNConv}{324{,}029\xspace}
\newcommand{\StwoNVisitors}{134{,}071\xspace}
\newcommand{\StwoNModels}{110\xspace}
\newcommand{\StwoNTurnsLater}{140{,}671\xspace}
\newcommand{\StwoNVotes}{80{,}205\xspace}
\newcommand{\StwoNRxAnswers}{659{,}969\xspace}
\newcommand{\StwoDurMedian}{20\xspace}
\newcommand{\StwoDurPTwoFive}{9.4\xspace}
\newcommand{\StwoDurPSevenFive}{39\xspace}
\newcommand{\StwoWaitMedian}{26\xspace}
\newcommand{\StwoSlowSd}{0.65\xspace}
\newcommand{\StwoContMean}{23.1\xspace}
\newcommand{\StwoContKMean}{51.2\xspace}
\newcommand{\StwoContRelOls}{8\xspace}
\newcommand{\StwoContDoubleOls}{1.3\xspace}
\newcommand{\StwoFsCoef}{0.94\xspace}
\newcommand{\StwoVoteIvOrderedDblHi}{+1.4\xspace}
\newcommand{\StwoVoteIvOrderedDblHiAbs}{1.4\xspace}
\newcommand{\StwoInstrSdRaw}{0.25\xspace}
\newcommand{\StwoVoteIvOrderedDblLo}{$-$5.4\xspace}
\newcommand{\StwoInstrSdExpPct}{10\xspace}
\newcommand{\StwoInstrSdCont}{$-$0.19\xspace}
\newcommand{\StwoRawBalLcmLogB}{+0.034\xspace}
\newcommand{\StwoRawBalLcmLogSE}{0.017\xspace}
\newcommand{\StwoInstrSdResid}{0.10\xspace}
\newcommand{\StwoSplineLoB}{+0.2\xspace}
\newcommand{\StwoSplineLoSE}{0.3\xspace}
\newcommand{\StwoSplineHiB}{$-$5.1\xspace}
\newcommand{\StwoSplineHiSE}{0.4\xspace}
\newcommand{\StwoDecNineAbs}{3.6\xspace}
\newcommand{\StwoDecNineSE}{0.5\xspace}
\newcommand{\StwoDecEightAbs}{0.9\xspace}
\newcommand{\StwoDecEightSE}{0.4\xspace}
\newcommand{\StwoRxpLikedOwnBx}{$-$0.26\xspace}
\newcommand{\StwoRxpLikedOwnSEx}{0.07\xspace}
\newcommand{\StwoRxpLikedPartnerBx}{+0.37\xspace}
\newcommand{\StwoRxpLikedPartnerSEx}{0.07\xspace}
\newcommand{\StwoRxpDislikedOwnBx}{+0.26\xspace}
\newcommand{\StwoRxpDislikedOwnSEx}{0.05\xspace}
\newcommand{\StwoTopicMin}{$-$8.7\xspace}
\newcommand{\StwoTopicMax}{+4.3\xspace}
\newcommand{\StwoTopicN}{10\xspace}
\newcommand{\StwoTopicCultureB}{+4.1\xspace}
\newcommand{\StwoTopicCultureSE}{2.0\xspace}
\newcommand{\StwoBinZeroWait}{10\xspace}
\newcommand{\StwoBinSevenWait}{40\xspace}
\newcommand{\StwoBinEightWait}{55\xspace}
\newcommand{\StwoBinNineWait}{111\xspace}
\newcommand{\StwoStart}{February 2025\xspace}
\newcommand{\StwoEnd}{May 2026\xspace}
\newcommand{\SampNConvAll}{410{,}016\xspace}
\newcommand{\SampNVotesAll}{149{,}209\xspace}
\newcommand{\SampNReactionsAll}{92{,}841\xspace}
\newcommand{\SampNStdMode}{376{,}662\xspace}
\newcommand{\SampNAutoIds}{119\xspace}
\newcommand{\SampNAutoConv}{14{,}893\xspace}
\newcommand{\SampNPooledConv}{50{,}628\xspace}
\newcommand{\SampNWithTiming}{345{,}809\xspace}
\newcommand{\SampNStdNonauto}{362{,}776\xspace}
\newcommand{\SampNAutoConvStd}{13{,}886\xspace}
\newcommand{\SampShareVoted}{36\xspace}
\newcommand{\SampShareReacted}{13\xspace}
\newcommand{\SampShareCont}{22\xspace}
\newcommand{\SthreeContOlsAbs}{1.5\xspace}
\newcommand{\SthreeContIvAbs}{3.5\xspace}
\newcommand{\SthreeContOlsHourAbs}{1.3\xspace}
\newcommand{\SthreeContOlsHourSE}{0.1\xspace}
\newcommand{\SthreeContIvHourAbs}{1.8\xspace}
\newcommand{\SthreeContIvHourSE}{0.7\xspace}
\newcommand{\SthreeContOlsIpAbs}{2.6\xspace}
\newcommand{\SthreeContIvIpAbs}{6.1\xspace}
\newcommand{\SthreeNReplies}{4{,}378{,}514\xspace}
\newcommand{\SthreeNConversations}{2{,}694{,}924\xspace}
\newcommand{\SthreeNIps}{1{,}829{,}326\xspace}
\newcommand{\SthreeLatMedian}{6\xspace}
\newcommand{\SthreeLatPTwoFive}{2\xspace}
\newcommand{\SthreeLatPSevenFive}{13\xspace}
\newcommand{\SthreeLatPNineFive}{43\xspace}
\newcommand{\SthreeContMean}{39.2\xspace}
\newcommand{\SthreeContMeanFirst}{17.6\xspace}
\newcommand{\SthreeLoadSd}{0.18\xspace}
\newcommand{\SthreeElasticity}{0.61\xspace}
\newcommand{\SthreeContMeanLater}{73.5\xspace}
\newcommand{\SthreeIvbalUsB}{$-$0.051\xspace}
\newcommand{\SthreeNHours}{19{,}077\xspace}
\newcommand{\SthreeBalhourUsB}{$-$0.023\xspace}
\newcommand{\SthreeBalhourUsSE}{0.010\xspace}
\newcommand{\SthreeDecOneAbs}{1.4\xspace}
\newcommand{\SthreeDecOneSE}{0.1\xspace}
\newcommand{\SthreeDecNineAbs}{2.2\xspace}
\newcommand{\SthreeDecNineSE}{0.2\xspace}
\newcommand{\SthreeAuditN}{5{,}167{,}182\xspace}
\newcommand{\SthreeAuditZero}{461{,}750\xspace}
\newcommand{\SthreeAuditOverSixZeroZero}{26\xspace}
\newcommand{\SthreeAuditNPrepared}{4{,}543{,}659\xspace}
\newcommand{\SthreeHcFsHourF}{10{,}068\xspace}
\newcommand{\SthreeHcContIvLaterHourB}{$-$3.6\xspace}
\newcommand{\SthreeHcContIvLaterHourSE}{0.9\xspace}
\newcommand{\SthreeHcContOlsHourIpAbs}{2.4\xspace}
\newcommand{\SthreeHcContOlsHourIpSE}{0.1\xspace}
\newcommand{\SthreeHcContIvHourIpAbs}{6.7\xspace}
\newcommand{\SthreeHcContIvHourIpSE}{1.2\xspace}
\newcommand{\SthreeHcContIvFirstHourB}{$-$1.2\xspace}
\newcommand{\SthreeHcContIvFirstHourSE}{1.1\xspace}
\newcommand{\SthreeHcContIvGptThreeFiveHourB}{$-$1.9\xspace}
\newcommand{\SthreeHcContIvGptThreeFiveHourSE}{0.2\xspace}
\newcommand{\SthreeHcContIvGptFourHourB}{+1.3\xspace}
\newcommand{\SthreeHcContIvGptFourHourSE}{1.4\xspace}
\newcommand{\SthreeHcContIvUsHourB}{+1.4\xspace}
\newcommand{\SthreeHcContIvUsHourSE}{1.8\xspace}
\newcommand{\SthreeHcGapOlsHourB}{+17.0\xspace}
\newcommand{\SthreeHcGapOlsHourSE}{0.7\xspace}
\newcommand{\SthreeHcContIvHourNozeroB}{$-$2.1\xspace}
\newcommand{\SthreeHcContIvHourNozeroSE}{0.7\xspace}
\newcommand{\SthreeHcInstrSdResidHour}{0.06\xspace}
\newcommand{\SthreeHcFsHourCoef}{0.72\xspace}
\newcommand{\SthreeHcShareMultimodel}{59\xspace}
\newcommand{\SthreeHcZeroShare}{8\xspace}
\newcommand{\SthreeStart}{April 2023\xspace}
\newcommand{\SthreeEnd}{July 2025\xspace}

%% file: sections/00_abstract.tex
\begin{abstract}
Answers from generative AI chatbots built on large language models (LLMs) now arrive at very different speeds:
reasoning models deliberate before replying, while response caches can return a stored answer at once. Blind
preference votes, which rank these models, are read as judgments of their text, and if delivery matters, faster answers
are expected to gain. Research on the labor illusion and on how text appears on screen suggests instead that an answer
which unfolds as it is written is credited over the same answer shown at once. We test these accounts with three sources
of variation in natural use. First, compar:IA, a French government chatbot arena, answered repeated prompts, by a random
draw, either with a fresh answer streamed as it was written or with a stored earlier answer from the same model, shown
at once. Within the same prompt, a replay lowered an answer's chance of winning a decisive blind vote by
\SoneVoteVThreeAbs percentage points. The same stored text won \SoneIdentEqwFresh\% of its decisive votes when streamed
and \SoneIdentEqwReplay\% when replayed, and voters called it less complete, more superficial and more often incorrect
when replayed. Second, provider congestion slowed streamed answers. In \StwoNConv conversations, slower answers earned no
detectable net credit in the vote, but a slower first exchange made visitors less likely to ask a follow-up question.
Third, in \SthreeNReplies WildChat replies from everyday ChatGPT use, replies slowed by congestion made users less likely
to continue. Delivery moved blind votes, in the direction opposite to the one the arena's designers expected, while
waiting cost follow-up questions that a comparison between two answers cannot register when both are slow. The results
bear on how AI services cache and stream answers, how they budget latency, and how preference leaderboards evaluate
models.
\end{abstract}

\keywords{response time \and latency \and generative AI \and large language models \and chatbots \and
streaming \and caching \and labor illusion \and operational transparency \and user engagement \and AI evaluation \and
human--AI interaction}

%% file: sections/01_intro.tex
\section{Introduction}
\label{sec:intro}

How long an AI system takes to answer has become a design choice. Reasoning models are built to spend more
computation, and therefore more time, before they reply \citep{guo2025deepseek,snell2024scaling}.
At the same time, engineers work to make answers arrive faster: speculative decoding shortens generation
\citep{leviathan2023speculative}, and response caches return a stored answer to a repeated question almost at once
\citep{bang2023gptcache}. Chatbot providers now offer ``fast'' and ``thinking'' modes side by side. Whether users
see an answer appear instantly, watch it being written, or wait while the system thinks is decided every day by
product teams, usually on cost grounds.

Blind preference votes are the main way generative AI models are compared. Public arenas show visitors two anonymous
answers to the same prompt and rank models by the votes \citep{chiang2024arena,termignon2026comparia}. The votes are read
as judgments of each model's text, and where delivery might matter, faster answers are expected to gain. The designers of
compar:IA, the French government's arena, write that ``Faster responses may be perceived as more fluent or reliable, while
slower responses may negatively impact user preference independently of content quality'' \citep{termignon2026comparia}.
People judge slowly produced algorithmic predictions as less accurate \citep{efendic2020slow}, responsiveness in LLM
applications ``is widely assumed to be critical'' \citep{tan2026latency}, and decades of research on system response
time link longer delays to lower satisfaction, weaker intentions to return, and less use
\citep{miller1968response,shneiderman1984response,guynes1988impact,palmer2002website,galletta2004delays,nah2004tolerable}.
A second reading treats time as a sign of deliberation and care. People rate products as better when they believe more
effort went into them \citep{kruger2004effort,morales2005effort}, and in a recent laboratory study, participants who
waited 2 seconds for a chatbot's answer to begin rated its answers as less thoughtful than participants who waited 9 or
20 seconds \citep{tan2026latency}.

Research on how output is presented gives reasons to expect the opposite of the speed reading for an answer that
appears at once instead of unfolding on screen. Displays of work raise the value people place on a service even when the
results are identical, an effect called the labor illusion \citep{buell2011labor}. Users rate a recommendation agent
higher when they perceive that it put more effort into its recommendations \citep{tsekouras2022scratch}, and beliefs
about the effort that went into a product help explain why consumers value AI-made products less
\citep{rix2025algorithm}. Research on waiting interfaces offers a second reason: displays that run during a wait hold
attention and immerse users \citep{lee2012online}, and generative AI responses presented word by word are accepted more
readily than responses presented line by line, because users feel more immersed \citep{jin2026word}. In an experiment
with AI writing tools, \citet{zhou2025thoughtful} expected the judged quality of a text not to vary with how it appeared
on screen; they found a small effect instead, with the highest ratings at a pace slightly faster than reading. Whether
such effects reach the blind choices that rank models, and in which direction, is not known.

Generative AI makes this question a routine operating decision. In the experiments that established the labor
illusion, the display of work was an animation shown while users waited for the result, such as a list of airlines
being searched while a travel site prepared its fares \citep{buell2011labor}. In a chatbot, the display is the answer
itself: streaming shows the text as it is written, and it is the default for nearly every answer. Operators can remove
it without changing a word. A response cache returns a stored answer that costs almost nothing to produce, and it can
show that answer at once or at any pace. Platforms already experiment with how generated output is revealed
\citep{jiang2026seeing}. If delivery moves blind votes, decisions that operators make for cost reasons, such as serving
stored answers, also change how good their answers seem, and with them the preference data used to rank models. This
paper asks how the delivery of the same answer, and the time an answer takes, change the votes that rank AI models and
whether users keep asking.

Settling these questions in the field is hard because response time is tied to content. Long answers take longer to
generate, difficult questions draw long answers, and reasoning models are slow and strong at once. A correlation
between waiting and satisfaction says little about what waiting does. Laboratory studies break that link, but they
assign tasks that participants must complete, so whether participants keep asking reflects the task as much as the
experience.

This paper uses three sources of variation in natural use that separate how long an answer takes from what it says.
The first is a lottery built into compar:IA, a public chatbot arena run by the French Ministry of Culture, in which
visitors put one prompt to two anonymous models and compare their answers side by side
\citep{termignon2026comparia}. From March 2026, when a model had already answered the same opening prompt, the
platform drew a random number. Depending on the draw, it either generated a new answer, which the visitor watched
being written, or replayed a stored earlier answer from the same model to the same prompt, which filled the screen in
a fraction of a second.
Replays reproduce earlier fresh answers word for word, and many stored texts appear in the data both as a fresh answer
and as a replay, so the comparison can also be made for the identical text. The second source is congestion. During our
study window the platform's server waited on each model provider in a way that let a slow provider hold up other
requests, and provider load varied from day to day. Answers of the same length from the same model therefore took
different times to arrive for reasons outside the visitor's control. The third source is WildChat, a corpus of several
million ChatGPT conversations whose logs record when each reply was created at the provider and when it was received
in full \citep{zhao2024wildchat}. There, too, provider congestion moves the wait for otherwise similar replies.

Three findings emerge. First, answers shown at once lost blind votes. Against a fresh answer, a replayed answer
won \SoneRawWinCached\% of decisive votes, those that picked one answer rather than declaring a tie. Comparing answers
to the same prompt, a replay lowered the chance of winning a decisive vote by \SoneVoteVThreeAbs percentage points
(standard error \SoneVoteVThreeSE), and among all eligible answers it lowered the share that won the conversation and
raised the share that lost it. The penalty persisted when
the same stored text was compared with itself. Averaged over \SoneIdentEqwN texts, the identical text won
\SoneIdentEqwFresh\% of its decisive votes when it was shown fresh and \SoneIdentEqwReplay\% when it was replayed,
and voters called it less complete and more often superficial and incorrect when it was replayed. Second, among
answers unfolding on screen, waiting earned no detectable net credit. An answer that took longer than usual for its length
did not win more often, and the estimates rule out gains of more than about \StwoVoteIvOrderedDblHiAbs points for a
doubling of the wait. Third, waiting has a behavioral cost. A slower first exchange lowered the chance that the visitor
asked a second question by \StwoContOlsAbs points per log unit of delay (about \StwoContRelOls\% of the base rate), or \StwoContDoubleOls points for each doubling of the wait. Most of the loss came in the slowest fifth of waits. In WildChat, replies slowed by congestion made users less likely
to continue, also when the comparison is limited to models used in the same clock hour.

These results run against both expectations about blind votes. Delivery moved them, in the direction opposite to the one the arena's designers anticipated. For the same text, a rapid replay instead of a fresh stream cut the share
of decisive wins roughly in half, and voters explained their choice in the language of substance. The time an answer
took did not move the vote detectably and showed up instead in whether users kept asking. Because a vote compares two
answers, a delay that slows both cannot register in it, although users still pay for it in follow-up questions they do
not ask.

The paper makes two theoretical contributions and one practical contribution. First, it shows in the field that how an
AI system reveals its output enters consequential judgments of that output with the wording held fixed. Research on the
labor illusion and on presentation has shown such effects on the value placed on a service and on laboratory ratings
\citep{buell2011labor,zhou2025thoughtful,jin2026word}. Here they move blind choices between models, by a wide margin, and in
the direction opposite to the reading of speed as fluency or competence
\citep{efendic2020slow,termignon2026comparia}. Second, it bounds what time adds once an answer is being written.
Slowdowns caused by congestion bought no detectable net credit in the vote, so the credit that laboratory evidence finds
for a later first word \citep{tan2026latency} does not extend to them, while they cost follow-up questions. Third, the
results bear on how AI systems are built and evaluated. Caching changes how answers are judged, latency changes whether
users keep asking, and preference leaderboards built from blind votes register how answers are delivered while missing
the cost of waiting.

The rest of the paper proceeds as follows. \Cref{sec:background} sets out the competing accounts and their
predictions. \Cref{sec:setting} describes compar:IA and WildChat. \Cref{sec:study1,sec:study2,sec:study3} present the
three studies, and \cref{sec:discussion} discusses implications for theory, design and evaluation.

%% file: sections/02_background.tex
\section{Theory and Hypotheses}
\label{sec:background}

How an AI service delivers an answer is a design choice. The operator can stream a fresh answer as the model writes it,
serve stored text at once, or let a slower model take its time. This section places that choice in recent research on
the design of AI agents, sets out the accounts that predict how the delivery of the same answer and the time an answer
takes affect judgments of it, and derives the competing predictions that the three studies test.

\subsection{The delivery of AI answers as a design choice}

Research in information systems and management has shown that the design cues of AI agents change how users evaluate
them and what users do. Human-like cues in a retailer's customer service chatbot changed whether customers accepted its
offers \citep{schanke2021humanizing}. Positive emotion expressed by an AI agent did less for service evaluations than
the same emotion from a human employee, in part because it violated what customers expected of an AI agent
\citep{han2023bots}. Response time is one such cue. In a laboratory experiment, a delayed chatbot response had opposing
effects on users' sense of social presence and on their usage intentions, and the effects differed between novice and
experienced users \citep{gnewuch2022opposing}. In a field experiment in customer service chats, customers who had just
seen a chatbot fail to understand them read unusually rapid replies from AI-assisted human agents as a sign that they
were still talking to a chatbot, and their sentiment fell \citep{zhang2026engaging}. Between people, fast replies signal
responsiveness and raise the chance of being hired \citep{hart2026speed}.

Generative AI adds a cue of its own: how the output is revealed. Platforms that generate content in seconds have begun
to experiment with how much of it to show and when, and in one field experiment revealing part of a generated output
before registration raised engagement more than revealing all of it \citep{jiang2026seeing}. These studies vary the source of an output, the agent's social
cues or how much of the output is shown, and the experiments closest to ours vary how the same text appears on screen
(\cref{sec:progressive}). What none observes is the same model's verbatim answer delivered in two ways under a real
operating policy and judged against another answer in a choice that ranks models.

\subsection{Speed and time as information about an answer}

A common account holds that people read the time an answer takes as information about the answer. It comes in two
versions with opposite signs. In the first, speed signals fluency and competence. The designers of compar:IA expect that
``Faster responses may be perceived as more fluent or reliable, while slower responses may negatively impact user
preference independently of content quality'' \citep{termignon2026comparia}. People judge slowly generated predictions
from algorithms as less accurate and rely on them less. Slowness signals effort for algorithms as for people, but because
prediction is seen as easy for an algorithm, the effort does not count toward quality \citep{efendic2020slow}. Practice
takes the same view: responsiveness in LLM applications ``is widely assumed to be critical'' \citep{tan2026latency}.
Research on human--computer interaction treated response time as a central design parameter from the start.
\citet{miller1968response} proposed thresholds for conversational transactions, and \citet{shneiderman1984response}
reviewed evidence that longer and more variable delays slow users down and frustrate them. Delays raise users' anxiety
\citep{guynes1988impact}, and the search for one acceptable response time has continued for decades
\citep{dabrowski2011forty}. Information systems research carried the question to the web. Download delay is one of the
performance metrics that predict a web site's success, including the likelihood that users return
\citep{palmer2002website}. In controlled experiments, longer page delays reduced users' performance, attitudes and
behavioral intentions \citep{galletta2004delays}, and users' tolerance for waiting is short \citep{nah2004tolerable}.
Search engines and large web sites report the same pattern: slower result pages change how people interact with them
\citep{arapakis2014latency}, and online experiments repeatedly show that speed matters for engagement
\citep{kohavi2014seven}.

In the second version, time signals deliberation and care. People rate a poem or a painting more highly when told that
it took longer to make \citep{kruger2004effort}, and observers judge others' decisions by how much thought went into
them, preferring the amount of thinking that fits the difficulty of the task \citep{kupor2014thought}. For large
language models, \citet{tan2026latency} varied the time before the first word of a streamed answer appeared.
Participants who waited 2 seconds rated the answers as less thoughtful than participants who waited 9 or 20 seconds,
and as less useful than those who waited 9 seconds; ratings after 9 and 20 seconds did not differ, and logged
interaction behavior did not change with the delay. The authors conclude that latency is ``a tunable design variable''.

\subsection{Progressive display}
\label{sec:progressive}

Modern chatbots stream each answer to the screen as the model produces it, so the user watches the text unfold at the
pace at which the machine writes it. We call this \emph{progressive display}. In a chatbot it differs from the displays
studied in services in two ways. The display is the answer itself rather than an animation beside it, so what unfolds
on the screen is the substance the user will judge. And it is the default: users see nearly every answer written out,
an answer that appears at once is the exception, and the operator decides when that happens, typically to avoid the cost
of generating the same answer again. Streaming exists to make answers start sooner, not to show the work, so we do not
treat it as operational transparency by design. The contrast our data observe is between a fresh answer streamed as it
was generated and a stored answer replayed in a fraction of a second.

Two lines of research predict that an answer which unfolds is judged better than the same answer shown at once. The
first runs through effort. Research on the labor illusion finds that people credit the work they can see
\citep{buell2011labor}. On simulated travel and dating web sites, users who watched a display of the work being done while
they waited valued the service more than users who received identical results at once. The value came from the display
rather than from the work: varying how many sites the service actually searched did not change how much users valued
it, while showing the search did. The display needed a wait, since there is nothing to show when results arrive at once,
and so the headline finding is that people ``can actually prefer websites with longer waits to those that return
instantaneous results'' \citep{buell2011labor}. Yet time added nothing once the display was present. When the same work
was shown over waits of 10 to 60 seconds, value was higher with the display at every wait and tended to fall as the wait
grew, with no interaction between the two. Consumers reward firms that visibly exert effort for them
\citep{morales2005effort}, and operational transparency can create value for customers and employees alike
\citep{buell2017reciprocal}. The logic also reaches machines: users rate a recommendation agent higher when they
perceive that it put more effort into its recommendations \citep{tsekouras2022scratch}, and beliefs about the effort
that went into a product help explain why consumers value AI-made products less \citep{rix2025algorithm}. Customers
also read the speed of a reply as evidence of how it was produced \citep{zhang2026engaging}. On this route, an answer shown at once looks as if
less work went into it, and people judge the quality of work partly by the effort they think it took
\citep{kruger2004effort}.

The second route runs through reading and attention. Interfaces that run during a wait hold attention and induce
focused immersion \citep{lee2012online}, and cues that convey progress reduce uncertainty while animated ones draw
attention away from the wait \citep{lee2017online}. Generative AI responses presented word by word are accepted more
readily than responses presented line by line, because users feel more immersed, although less so when users need to
understand the content \citep{jin2026word}. In an experiment with AI writing tools that assigned the same texts to
different presentation speeds, \citet{zhou2025thoughtful} expected that ``There will be no variance between the judged
quality of the text''. They found a small effect instead. The pace rated highest for quality, slightly faster than
average reading speed, was also the one rated most comfortable to read. On this route, text that unfolds at a readable
pace is easier to follow, and a block that appears at once is skimmed or looks like a stock answer.

The two routes differ in what the display must do: the first needs it to be taken as a sign of the work behind this
answer, the second only needs the text to be easier to follow. Both predict that the answer shown at once is judged
worse. Our data record votes, the qualifiers voters attach to them, reactions and later actions, so they test whether
progressive display is credited, not which route carries the credit; \cref{sec:limits} describes a test that would
separate them. Both routes also predict that extra time adds nothing once the answer is unfolding: the displayed work is
the same text whether it takes longer or not, and a pace slower than reading makes the text no easier to follow.

\subsection{Competing predictions}

\Cref{tab:predictions} sets out what the accounts predict. The first observable is how the same answer fares when it is
shown at once rather than unfolding. If votes judge only the text, as \citet{zhou2025thoughtful} expected for ratings,
the two should do equally well. If speed signals fluency or competence, as compar:IA's designers expected, the answer
shown at once should do at least as well. If progressive display is credited, it should do worse.

\begin{quote}
\textbf{H1a (progressive display).} An AI answer shown at once is judged worse than a comparable answer that unfolds on
screen as it is generated.\\
\textbf{H1b (text only, or speed as fluency).} An AI answer shown at once is judged no worse than a comparable answer
that unfolds on screen as it is generated.
\end{quote}

The qualifiers that voters can attach to a vote, such as complete, superficial or incorrect, show how they explain their
choice. We use them to describe the judgment, not to tell the routes apart.

The second observable is the time taken by answers that unfold on screen. Here the two readings of time part ways, and
progressive display predicts no change.

\begin{quote}
\textbf{H2a (time as deliberation).} Among answers that unfold on screen, an answer that takes longer is judged better.\\
\textbf{H2b (time as a cost).} Among answers that unfold on screen, an answer that takes longer is judged worse.
\end{quote}

A result between the two would fit progressive display. It would also arise if a credit for deliberation and a cost of waiting cancelled, so the second observable bounds the net effect of time on the vote rather than identifying
its parts.

The third observable is behavior. A chatbot conversation is a sequence of service encounters: after each answer the
user decides whether to start another one. Research on queues finds that delays lower customers' evaluations of a
service, largely through anger and uncertainty \citep{taylor1994waiting}, and that longer waits lead people to abandon
service, whether patients in an emergency department \citep{batt2015waiting} or callers to a call center
\citep{aksin2013structural}. If time is a cost, a costly wait for one answer makes the next encounter less attractive.

\begin{quote}
\textbf{H3 (cost of time).} A longer wait for an AI answer lowers the chance that the user continues the conversation.
\end{quote}

H3 does not tell the accounts of H1 apart. It locates where the cost of time appears, and it matters for evaluation:
a vote compares two answers, so a delay that slows both cannot change which one wins, while the decision to ask again
registers it.

\begin{table}[t]
\centering
\caption{Competing predictions.}
\label{tab:predictions}
\small
\begin{tabularx}{\linewidth}{>{\raggedright\arraybackslash}p{0.3\linewidth}>{\raggedright\arraybackslash}X}
\toprule
Observable (study) & Prediction, by account \\
\midrule
The same answer shown at once instead of unfolding (Study 1) & Text only: no difference. Speed as fluency or competence: the answer shown at once does at least as well. Progressive display, through effort or reading: it does worse (H1a against H1b) \\
\addlinespace
An answer that unfolds takes longer (Study 2) & Time as deliberation: it wins more often (H2a). Time as a cost: it wins less often (H2b). Progressive display: no change \\
\addlinespace
The user waited longer for an answer (Studies 2 and 3) & Cost of waiting: fewer follow-up questions (H3) \\
\bottomrule
\end{tabularx}
\end{table}

\subsection{Preferences as the yardstick for AI systems}

These questions also bear on how generative AI is evaluated. Public arenas show visitors two anonymous answers to the
same prompt and aggregate their votes into leaderboards \citep{chiang2024arena,termignon2026comparia}. Human
preferences of this kind respond to features other than accuracy, most prominently answer length
\citep{singhal2024long}, and, as noted above, the designers of compar:IA expect latency to influence judgments too
\citep{termignon2026comparia}. Whether such influences are a problem depends on what a leaderboard is meant to
measure. If it measures the quality of the text a model writes, how the text is delivered is a confound. If it
measures the service a user experiences, delivery belongs in the score, but then the cost of a slow model should
appear in it, and a vote cast after the wait may not register the follow-up questions the wait has cost.

%% file: sections/03_setting.tex
\section{Setting and Data}
\label{sec:setting}

\subsection{compar:IA}

compar:IA is a public service developed inside the French government to let the general public compare
conversational AI models and to collect French-language preference data \citep{termignon2026comparia}. No account is
needed, and the platform is widely used in schools and universities. A visitor types a prompt, or picks a suggested
one, and chooses how the two models are drawn: at random, from a list the visitor selects, or within themes such as
large versus small models or reasoning models. Two anonymous models then answer side by side. The two answers appear together once both models have begun to
answer. Each is then shown on screen as its text arrives, with a loading indicator until the model finishes, and for
reasoning models the reasoning text appears in a panel above the answer.\footnote{We read the interface and the
server's behavior from the platform's public source code and its commit history \citep{comparia2026code}.
\Cref{app:implementation} lists the code changes that matter for our designs, with links to them and their merge
dates.} The visitor can then ask a follow-up question, which goes to both
models, or react to any individual answer with a thumbs-up or thumbs-down and optional qualifiers (useful, complete,
creative, clear formatting; incorrect, superficial, instructions not followed). If no message reaction was given, the
visitor can also cast a final vote for one model or declare the two equal, again with optional qualifiers. The models' identities are
revealed only after this feedback.

The platform publishes three datasets under the open Etalab 2.0 license \citep{comparia2026data}: conversations,
final votes and message reactions. We use the release that runs to 18 May 2026. It contains \SampNConvAll
conversations from October 2024 onward, \SampNVotesAll votes and \SampNReactionsAll reactions, with an anonymized
visitor identifier and, for each completed answer, its text, its reasoning text and metadata. From February 2025, the metadata
record each answer's \emph{duration}, measured on the platform's server: the time from the moment it sent the request
to the model provider to the moment the last piece of the answer arrived. It includes both the wait for the first
words and the time spent writing, but not the time needed to carry the text to the visitor's browser. The metadata
also flag answers served from the platform's cache, which we describe in \cref{sec:study1}. 

We keep conversations in the platform's five standard modes (\SampNStdMode conversations) and remove automated
traffic: \SampNAutoIds visitor identifiers that started more than 20 conversations on a single day account for
\SampNAutoConv conversations and almost never vote. One further identifier is shared by \SampNPooledConv
conversations from visitors without a stored cookie; we keep these conversations, treat each of their sessions as a
separate visitor for clustering, and leave them out of every within-visitor comparison. In the remaining data, the
visitor asked a second question in \SampShareCont\% of conversations, cast a final vote in \SampShareVoted\%, and
reacted to at least one message in \SampShareReacted\%. \Cref{tab:flow} in the appendix traces each analysis sample
from the release.

Two features of how the platform ran during our window matter for the design, and both are documented in its
source code \citep{comparia2026code}. First, from March 2026 the platform kept a cache of first answers, which
\cref{sec:study1} exploits. Second, until October 2026 the server waited for model providers in a way that held up
other requests. The developers' later fix describes the problem: the answer stream ran inside the request handler
``so every wait on the provider held the one event loop. While a model was slow to answer, every other visitor's
request waited too, and so did the health check.'' Generation time therefore varied with load on the platform and on
each provider, not only with the question and the answer, which \cref{sec:study2} uses.

\subsection{WildChat}

WildChat offered free access to ChatGPT through a public web application in exchange for users' consent to the
anonymous collection of their chat transcripts \citep{zhao2024wildchat}. We use its largest public release,
WildChat-4.8M \citep{wildchat2025data}, which covers conversations from April 2023 to July 2025 with several
generations of OpenAI models (GPT-3.5, GPT-4, GPT-4o, GPT-4o-mini, GPT-4.1-mini and o1). The public release
contains the conversations that were not flagged as toxic. For every reply, the logs record the time, to the second,
at which the provider created the completion, and the time at which WildChat's server received the full reply,
together with a hashed IP address. The difference runs from the creation time that the provider assigned to the completion to
the receipt of the complete reply, so it covers the production and delivery of the reply after that timestamp. WildChat users chatted with one model at a time and
gave no ratings, so this source speaks to continuation only.

%% file: sections/04_study1.tex
\section{Study 1: Answers That Appear at Once}
\label{sec:study1}

\subsection{The replay lottery}

From March 2026 the platform kept a cache, a store of earlier first answers, to reduce its costs \citep{comparia2026code}. The store
was keyed by the model and the text of the opening prompt, with surrounding spaces removed and capital letters
ignored; it held up to the five most recent answers for each key and expired 48 hours after the key was last used.
For each of the two answers in a conversation, the platform drew a random number. With a fixed probability it looked
for the key in the store and, if the key was there, replayed one of the stored answers, chosen at random; otherwise it
asked the model for a fresh answer, which was then added to the store. The first replay in the data occurs on
\SoneStart, and our window runs to \SoneEnd. \Cref{app:implementation} documents the code that ran in this window.

\begin{figure}[t]
\centering
\includegraphics[width=0.8\linewidth]{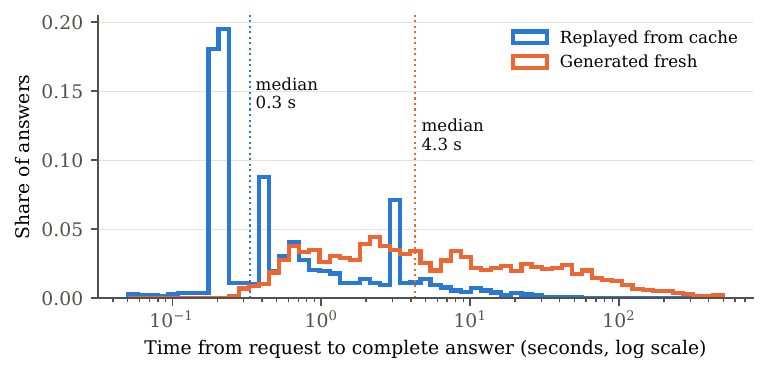}
\caption{Time from request to complete answer for eligible first answers in compar:IA, \SoneStart to \SoneEnd,
measured on the platform's server. Replays were served from the platform's store of earlier answers; fresh answers
were generated and written on screen. Dotted lines mark the medians.}
\label{fig:durations}
\end{figure}

A fresh answer was written on screen at the pace at which the model produced it. A replay was meant to look similar:
the code breaks the stored text into about fifteen pieces and ``yields with small delays to mimic real streaming
behavior for consistent UX'' \citep{comparia2026code}. In practice the delays were cut short. The server read the two
answers of a comparison together and, as the developers later explained, ``the slower read was cancelled as soon as
the other model answered,'' which ended the replay's pause each time the other answer produced text. The recorded
durations, which the server measured from the start of the replay to its last piece, show the result
(\cref{fig:durations}). The median replay took \SoneDurCachedMedian seconds, and three quarters took at most
\SoneDurCachedPSevenFive seconds. The median fresh answer to the same set of prompts took \SoneDurFreshMedian seconds
(interquartile range \SoneDurFreshPTwoFive to \SoneDurFreshPSevenFive seconds). Within the same model and prompt, a
replay shortened the recorded duration by \SoneReplayLdurAbsLog log units, a factor of more than five. A replayed
answer therefore filled its side of the screen in a fraction of a second, far faster than a reader can follow text
being written, while the answer beside it was usually still being written. For reasoning models, the stored reasoning
text appeared in one piece. We call this treatment a replay and compare it with fresh generation.

The public data flag which answers were replays but not which answers could have been. We reconstruct eligibility
from the conversation logs with the platform's own matching rule, which compares prompts after removing surrounding
spaces and ignoring capital letters. An answer is eligible if the same model had already given a fresh answer to the
same opening prompt during the caching period, and if that model--prompt key had been used within the previous 48
hours. This
yields \SoneNEligAnswers eligible answers in \SoneNEligCells model--prompt cells (\SoneNEligPrompts distinct
prompts, \SoneNEligVisitors identified visitors), of which \SoneShareCachedElig\% were replays. The replay share was
\SoneShareCachedRegimeOne\% until 5 April and \SoneShareCachedRegimeTwo\% afterwards, consistent with a change in the
configured probability, so every comparison holds the week fixed. Many eligible prompts are short questions that many visitors asked in identical form, such as
asking for the capital of France. Others are longer exercises repeated word for word, such as summarizing a given text.
The most frequent prompt accounts for \SoneTopPromptShare\% of eligible
answers and the ten most frequent for \SoneTopOneZeroPromptShare\%. The reconstruction misses some eligible answers,
because the public release omits some conversations: \SoneAuditIneligShare\% of all replays in the window
(\SoneAuditCachedIneligNorm of \SoneAuditNCached) occur where the logs show no earlier fresh answer for the key.
Matching prompts by their exact text instead of the platform's rule would have missed a further \SoneAuditNEligNormOnly
eligible answers.

\subsection{Is the draw as good as random?}

Three checks support treating the draw as random within a model--prompt cell. First, replays are what the code says
they are. Of all replays in the window, \SoneReplayMatchShare\% reproduce word for word an earlier fresh answer from
the same model to the same prompt; the rest most likely reproduce answers from conversations that are not in the public
release. Second, the measured content of replayed and fresh answers is similar. Within model--prompt cells, replayed
answers are \SoneEquivLchAbs\% longer in characters (standard error \SoneEquivLchSE), \SoneEquivLtokAbs\% longer in
tokens (\SoneEquivLtokSE) and carry \SoneEquivLreasAbs\% more reasoning text (\SoneEquivLreasSE). Third, the draw is
unrelated to the visitor and the context (\cref{tab:s1balance}). Comparing replayed and fresh answers within the same
week only, some characteristics differ, because the measured replay share varied across prompts and models; within
model--prompt cells, every difference is close to zero. The main specification therefore compares answers to the same
prompt.

\input{generated/tab_s1_balance}

\subsection{Estimation}

The main outcome is the final vote. For conversation $i$, let $W_i=1$ if the left-hand answer won a decisive vote,
$R_{i\ell}$ and $R_{ir}$ indicate whether the left and right answers were replays, and $E_{i\ell}$ and $E_{ir}$
whether they were eligible. We estimate
\begin{equation}
W_i = \beta\,(R_{i\ell}-R_{ir}) + \alpha_{(E_{i\ell}-E_{ir}),\,t(i)} + \gamma_{p(i)} + \delta_{q(i)} + \varepsilon_i ,
\label{eq:vote}
\end{equation}
where $\alpha$ is a fixed effect for the eligibility pattern in week $t$, $\gamma$ a fixed effect for the unordered
pair of models, and $\delta$ a fixed effect for the prompt. The left and right positions are assigned at random, so
the two sides are symmetric, and $\beta$ is the change in an answer's probability of winning a decisive vote when it
is replayed rather than generated. Standard errors are clustered by visitor. The sample contains the
\SoneNVotesDecisive decisive votes in conversations with at least one eligible answer. A replay could change whether a
visitor votes and whether the vote is a tie. We therefore also report outcomes that do not condition on a decisive
vote: how every conversation with an eligible answer ended, a score that counts a tie as half a win, and bounds that
allow for selection into voting.

\subsection{Results}

\begin{figure}[t]
\centering
\includegraphics[width=\linewidth]{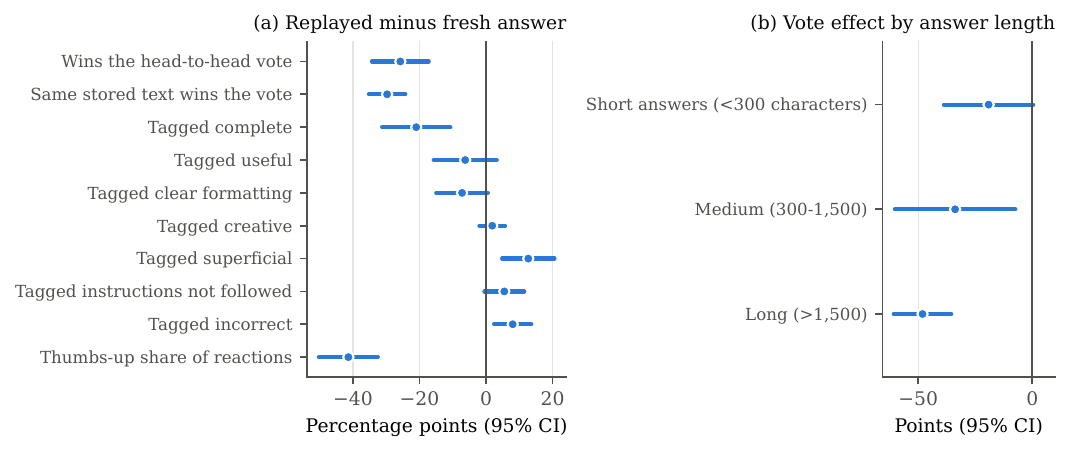}
\caption{Study 1: how a replayed answer is judged, relative to a fresh answer to the same prompt. Panel (a): the vote,
the vote for the identical stored text, the qualifiers attached to the vote (left-minus-right difference) and the
share of thumbs-up among reactions. Panel (b): the vote effect by the mean length of the two answers. Bars are 95\%
confidence intervals from standard errors clustered by visitor.}
\label{fig:cache}
\end{figure}

\paragraph{Replayed answers lose.} In the \SoneNVotesOnecached decisive votes between a replayed and a fresh answer,
the replayed answer won \SoneRawWinCached\% of the time. \Cref{tab:s1votes} shows that the penalty is stable across
specifications. A replay lowers the chance of winning a decisive vote by \SoneVoteVOneAbs points with
eligibility-by-week fixed effects, \SoneVoteVTwoAbs with model-pair fixed effects and \SoneVoteVThreeAbs with prompt
fixed effects, our main specification. With mode fixed effects and controls for the length and reasoning length of both
answers, it is \SoneVoteVFourAbs points. Among conversations in which both answers were eligible, so that both sides were
drawn at random, the estimate is \SoneVoteBotheligPhAbs points (\SoneVoteBotheligPhSE). In those conversations, the
left answer won \SoneStateFFWinA\% of decisive votes when both answers were fresh and \SoneStateRRWinA\% when both were
replays, and the replay won \SoneStateDiscordantReplayWin\% when one was replayed and the other fresh. Clustering the
standard errors by prompt as well as by visitor, to allow for repeated prompts, changes them little. Supporting H1a over H1b,
an answer shown at once is judged worse than an answer that unfolds on screen.

\input{generated/tab_s1_votes}

\paragraph{The identical text.} A replay reproduces an answer that the same model gave earlier to the same prompt, and
that earlier answer was itself shown fresh to another visitor. The data therefore contain repeated appearances of the
same stored text, some written on screen and some replayed. \SoneIdentNTexts stored texts appear in both forms, and
\SoneIdentDecisiveBothTexts of them received decisive votes in both forms: \SoneIdentdNBothFresh fresh and
\SoneIdentdNBothReplay replayed appearances, judged by \SoneIdentdNBothVis visitors across \SoneIdentdNBothPrompts
prompts. The texts match in the answer itself, including its formatting. For the \SoneIdentReasAnyShare\% of texts that
came with reasoning, the reasoning had the same length in every appearance in \SoneIdentReasSameShare\% of cases. A
replay came a median of \SoneIdentdHSinceMedian hours after the fresh appearance of its text. Weighting each text
equally, the same text won \SoneIdentEqwFresh\% of its decisive votes when it was shown fresh and
\SoneIdentEqwReplay\% when it was replayed, a difference of \SoneIdentEqwAbs points (\SoneIdentEqwSE). With fixed
effects for the text, a replay lowered its chance of winning by \SoneIdentDecisiveAbs points (\SoneIdentDecisiveSE;
\cref{tab:s1estimands}c). The estimate is similar with standard errors clustered by text (\SoneIdentDecisiveCltextSE) and with fixed
effects for the opponent's model and a control for the length of its answer (\SoneIdentDecisiveOppAbs points). It is
also similar when replays are limited to the first 48 hours after the fresh appearance (\SoneIdentDecisiveFourEighthAbs
points) and among the \SoneIdentDecisiveEligTexts texts whose fresh and replayed appearances were all eligible for the
draw (\SoneIdentDecisiveEligAbs points; \SoneIdentDecisiveEligCltextSE clustered by text). The identical text also drew
fewer thumbs-up and more thumbs-down when it was replayed. This comparison holds the words fixed rather
than drawing the delivery at random, so it complements the lottery: it shows that differences between texts do not
account for the penalty.

\paragraph{How the conversations ended.} \Cref{tab:s1estimands}a follows every eligible answer recorded in the release to
the end of its conversation, without conditioning on a vote. A replay did not change whether the conversation ended in a final vote
(\SoneJointONovoteB points, \SoneJointONovoteSE, from \SoneJointONovoteBase\% without a vote) or in a tie
(\SoneJointOTieB, \SoneJointOTieSE). It lowered the share of conversations that the answer won by
\SoneJointOWinAbs points and raised the share it lost by \SoneJointOLoseAbs points, from bases of \SoneJointOWinBase\% and \SoneJointOLoseBase\%. Among
voted answers, a score that counts a tie as half a win falls by \SoneJointOScoreAbs points (\SoneJointOScoreSE).
Lee bounds allow a replay to change who votes, under the assumption that it moves every visitor's voting in the
same direction \citep{lee2009training}. They put the effect on the decisive-vote win share between \SoneLeeDecLower and
\SoneLeeDecUpper points (95\% interval \SoneLeeDecCiLo to \SoneLeeDecCiHi), and the effect on the tie-inclusive score
between \SoneLeeAnyLower and \SoneLeeAnyUpper points (\SoneLeeAnyCiLo to \SoneLeeAnyCiHi). Computed separately for the
two periods of the replay probability, the bounds on the decisive-vote win share are \SoneLeeDecROneLower to
\SoneLeeDecROneUpper and \SoneLeeDecRTwoLower to \SoneLeeDecRTwoUpper points. A replay on one side also
made a tie less likely among voters, by \SoneTieAbsdcAbs points (\SoneTieAbsdcSE) from a base of \SoneTieAbsdcMean\%:
voters saw a difference between a replay and a fresh answer and resolved it against the replay.

\input{generated/tab_s1_estimands}

\paragraph{What voters say.} The qualifiers attached to votes describe the replayed answer as worse on substance
(\cref{tab:s1judg} and \cref{fig:cache}a). Relative to the fresh answer, the replay was \SoneTagCompleteAbs points
(\SoneTagCompleteSE) less likely to be called complete, \SoneTagSuperficialAbs points (\SoneTagSuperficialSE) more
likely to be called superficial and \SoneTagIncorrectAbs points (\SoneTagIncorrectSE) more likely to be called
incorrect. The differences for usefulness, clear formatting and following instructions point the same way but are less
precise. The same descriptions follow the identical text: with fixed effects for the text, a replay was
\SoneIdentTagCompleteAbs points (\SoneIdentTagCompleteSE) less likely to be called complete,
\SoneIdentTagSuperficialAbs points (\SoneIdentTagSuperficialSE) more likely to be called superficial and
\SoneIdentTagIncorrectAbs points (\SoneIdentTagIncorrectSE) more likely to be called incorrect (last column of
\cref{tab:s1judg}). Voters thus explained a choice against the same words, when they appeared at once, in the language of substance:
incomplete, superficial, incorrect. The qualifiers are optional and chosen with the vote, so they show how voters
justify their choice rather than what they believe about a particular claim. Even so, they run against the reading of
speed as fluency or competence, in which an answer that arrives at once should seem at least as reliable.
Thumbs on individual answers tell the same story. A replayed answer received \SoneRxLikedRel\% fewer
thumbs-up and \SoneRxDislikedRel\% more thumbs-down than a fresh answer to the same prompt from the same model, and
among the answers that drew any reaction, the share of thumbs-up was \SoneRxLikeshareAbs points lower
(\SoneRxLikeshareMpAbs points within model--prompt cells).

\input{generated/tab_s1_judgments}

\paragraph{Absolute and relative penalties.} Votes compare two answers, so a replay may lose partly because the answer
beside it was visibly being written. Thumbs reactions allow a check, because they attach to one answer and because
some conversations had two replays. The reference group is answers in conversations where both answers were fresh,
which drew a thumbs-up \SoneRxCellbaseLiked\% of the time and a thumbs-down \SoneRxCellbaseDisliked\%. Relative to
that group, and within model--prompt cells, a replayed answer received \SoneRxCellLikedCOneZeroAbs points fewer
thumbs-up when the other answer was fresh (\SoneRxCellLikedCOneZeroSE) and \SoneRxCellLikedCOneOneAbs points fewer when
the other answer was also a replay (\SoneRxCellLikedCOneOneSE). The thumbs-up rate of a fresh answer beside a replay
changed by \SoneRxCellLikedCZeroOneB points (\SoneRxCellLikedCZeroOneSE). Thumbs-down rose by \SoneRxCellDislikedCOneZeroAbs
(\SoneRxCellDislikedCOneZeroSE) and \SoneRxCellDislikedCOneOneAbs points (\SoneRxCellDislikedCOneOneSE) for the replay in
the two cases and changed by \SoneRxCellDislikedCZeroOneB points (\SoneRxCellDislikedCZeroOneSE) for a fresh answer
beside a replay. Replays are thus judged worse even when nothing on the screen is being written, and the penalty is larger
when a fresh answer is being written beside them: in a model with the answer's own replay, its neighbor's replay and
their interaction, the interaction is \SoneRxintLikedXIntB points for thumbs-up (\SoneRxintLikedXIntSE).

\paragraph{The penalty across answer lengths.} The penalty appears for short, medium and long answers
(\cref{fig:cache}b and \cref{tab:s1robust}): \SoneHetShortAbs points for answers under 300 characters on average,
\SoneHetMediumAbs points for medium answers and \SoneHetLongAbs points for answers longer than 1,500 characters. A linear
interaction with the expected length of the eligible answer, the mean log length of the earlier fresh answers from the
same model to the same prompt, is \SoneLenPriorB points per log unit (\SoneLenPriorSE). The penalty appears whether or not a reasoning model was in the pair (\SoneHetReasoningAbs and
\SoneHetNoreasoningAbs points).

\input{generated/tab_s1_robust}

\paragraph{Robustness.} A stored answer can be out of date when the question depends on the date, as with a request
for today's date. Excluding prompts that refer to time or current events leaves the estimate at \SoneRobTimesensAbs
points. The estimate is \SoneRobNotopAbs points without the prompt that drew the most votes
(\SoneTopVotePromptShare\% of the vote sample) and \SoneRobPopularAbs points for prompts asked by at least three
visitors. It is \SoneRobNopooledAbs points without the shared cookie-less identifier, and \SoneRobVerifiedAbs points
when every replay in the sample is a word-for-word match to an earlier fresh answer. The penalty is present in both periods of the replay probability
(\SoneRobRegimeOneAbs and \SoneRobRegimeTwoAbs points). Stricter definitions of eligibility leave it about the same:
\SoneRobWTwoFourAbs points (\SoneRobWTwoFourSE) when the key must have been used within the previous 24 hours. In the
smaller sample in which the store had served a replay for the key within the previous 48 hours, so that it was
demonstrably in use, the estimate is \SoneRobHitAbs points (\SoneRobHitSE). Serving one answer from the store also shortened the
recorded time of the fresh answer beside it, by about \SoneSpillPct\%. As an illustration, applying the preferred vote
estimate of Study 2 and its 95\% interval to a change of that size implies a change of \SoneSpillImpliedLo to
\SoneSpillImpliedHi points in the replay's chance of winning.

\paragraph{Continuation.} Replays did not detectably change whether visitors asked a second question: each replayed
first answer changed that probability by \SoneEngContB points (\SoneEngContSE; 95\% interval \SoneEngContCILo to
\SoneEngContCIHi) from a base of \SoneEngContMean\% (\cref{tab:s1engage} in the appendix). The eligible prompts are short and often assigned, so few of these visitors continue; the congestion study that
follows is designed to measure continuation.

%% file: generated/tab_s1_balance.tex
\begin{table}[t]
\centering
\caption{Study 1: within a model and prompt, the replay draw is unrelated to who asked and what was answered.}
\label{tab:s1balance}
\small
\begin{tabular}{lccc}
\toprule
 & Mean & \multicolumn{2}{c}{Replayed minus fresh} \\
\cmidrule(lr){3-4}
 & & Week FE & Week + model$\times$prompt FE \\
\midrule
\multicolumn{4}{l}{\emph{Pre-determined characteristics}} \\
Hour of day (0--23) & 11.87 & 0.085 (0.112) & 0.094 (0.132) \\
Day of week (0 = Monday) & 2.72 & 0.038 (0.037) & $-$0.011 (0.036) \\
Visitor's earlier conversations (log) & 0.82 & 0.034 (0.026) & 0.028 (0.017) \\
Prompt length (log characters) & 3.78 & $-$0.052 (0.027) & constant in cell \\
Visitor chose the models (share) & 0.81 & 0.055 (0.008) & 0.003 (0.003) \\
Model size (log parameters) & 5.21 & $-$0.147 (0.031) & constant in cell \\
Science and technology topic (share) & 0.14 & $-$0.024 (0.007) & 0.000 (0.003) \\
Shown in the right-hand position (share) & 0.50 & $-$0.006 (0.010) & $-$0.007 (0.011) \\
\midrule
\multicolumn{4}{l}{\emph{Content of the answer}} \\
Answer length (log characters) & 4.13 & & 0.013 (0.007) \\
Output tokens (log) & 3.97 & & 0.028 (0.010) \\
Reasoning text (log characters) & 2.04 & & 0.047 (0.017) \\
\bottomrule
\end{tabular}
\par\smallskip
\parbox{0.95\linewidth}{\footnotesize Notes: 11{,}699 eligible first-turn answers
(\SoneShareCachedElig\% replayed). Each cell is the coefficient on a replay indicator (standard error clustered by
visitor). Prompts are matched as the platform matched them, after removing surrounding spaces and ignoring capital
letters. Model$\times$prompt cells are the units within which the platform drew between replay and fresh generation;
prompt length and model size do not vary within a cell. The visitor-history row excludes the shared identifier used
for visitors without a cookie.}
\end{table}

%% file: generated/tab_s1_votes.tex
\begin{table}[t]
\centering
\caption{Study 1: a replayed answer loses the blind vote. Dependent variable: the left-hand answer wins a decisive vote (percentage points).}
\label{tab:s1votes}
\small
\begin{tabular}{lccccc}
\toprule
 & (1) & (2) & (3) & (4) & (5) \\
\midrule
Replayed (left) minus replayed (right) & $-$26.2 (2.7) & $-$23.7 (3.2) & $-$25.8 (4.4) & $-$25.7 (4.3) & $-$30.3 (5.6) \\
\quad SE clustered by visitor and prompt &  & (2.9) & (4.8) &  &  \\
\midrule
Eligibility $\times$ week FE & Yes & Yes & Yes & Yes & \\
Week FE & & & & & Yes \\
Model-pair FE & & Yes & Yes & Yes & Yes \\
Prompt FE & & & Yes & Yes & Yes \\
Mode FE, length and reasoning controls & & & & Yes & \\
Sample & \multicolumn{4}{c}{At least one side eligible} & Both sides eligible \\
Decisive votes & 1{,}005 & 1{,}005 & 1{,}005 & 1{,}005 & 363 \\
Clusters (visitors) & 862 & 862 & 862 & 862 & 300 \\
\bottomrule
\end{tabular}
\par\smallskip
\parbox{0.95\linewidth}{\footnotesize Notes: Column (3) is the main specification. Votes that declared the two
answers equal are excluded here; Table~\ref{tab:s1estimands} reports outcomes that include them. The regressor is $+1$
when only the left answer was replayed, $-1$ when only the right answer was replayed and 0 otherwise. Standard errors
clustered by visitor in parentheses.}
\end{table}

%% file: generated/tab_s1_estimands.tex
\begin{table}[t]
\centering
\caption{Study 1: the replay penalty without conditioning on a decisive vote, and for identical text (percentage points).}
\label{tab:s1estimands}
\small
\begin{tabular}{lcc}
\toprule
 & Mean for fresh (\%) & Replayed minus fresh \\
\midrule
\multicolumn{3}{l}{\emph{(a) Every eligible answer: how the conversation ended (11{,}699 answers)}} \\
\quad Own answer wins a decisive vote & 11.2 & $-$1.5 (0.4) \\
\quad Other answer wins a decisive vote & 11.0 & 2.3 (0.4) \\
\quad Vote declares the two equal & 15.3 & $-$0.5 (0.4) \\
\quad No final vote & 62.5 & $-$0.3 (0.6) \\
\quad Net preference (wins minus losses) & 0.2 & $-$3.8 (0.6) \\
\midrule
\multicolumn{3}{l}{\emph{(b) Answers that received a final vote, and bounds for selection into voting}} \\
\quad Score counting a tie as half a win (all 2{,}180 voted answers) & 50.3 & $-$15.9 (2.3) \\
\quad Lee bounds, decisive-vote win share & & [$-$39.6, $-$28.9] \\
\quad \quad 95\% interval & & [$-$52.4, $-$20.3] \\
\quad Lee bounds, score among all votes & & [$-$28.6, $-$17.0] \\
\quad \quad 95\% interval & & [$-$36.0, $-$8.9] \\
\quad Lee bounds, decisive-vote win share, 16 March--5 April & & [$-$41.8, $-$22.9] \\
\quad Lee bounds, decisive-vote win share, 6 April--18 May & & [$-$37.8, $-$33.2] \\
\midrule
\multicolumn{3}{l}{\emph{(c) The same stored text, shown fresh and replayed (265 texts with decisive votes in both forms)}} \\
\quad Wins a decisive vote, each text weighted equally & 56.8 & $-$27.3 (3.7) \\
\quad Wins a decisive vote, text fixed effects & 55.0 & $-$29.8 (2.8) \\
\quad \quad standard errors clustered by text & & (3.5) \\
\quad \quad adding week fixed effects & & $-$29.8 (2.8) \\
\quad \quad adding the other model and the length of its answer & & $-$30.2 (2.9) \\
\quad \quad replays within 48 hours of the fresh appearance & & $-$28.0 (3.0) \\
\quad \quad only eligible appearances (44 texts) & & $-$28.2 (6.4) \\
\quad \quad \quad clustered by text & & (9.1) \\
\quad Thumbs up on the answer, text fixed effects & & $-$1.33 (0.22) \\
\quad Thumbs down on the answer, text fixed effects & & 1.10 (0.19) \\
\bottomrule
\end{tabular}
\par\smallskip
\parbox{0.95\linewidth}{\footnotesize Notes: Panels (a) and (b): answer-level regressions on the answer's own replay
indicator and the other answer's replay indicator, with model$\times$prompt fixed effects and week fixed effects
interacted with the other answer's eligibility; the fresh mean is for conversations in which neither answer was
replayed. Lee bounds use conversations with exactly one eligible answer (2{,}097); they trim the
group in which a replay made voting more (decisive votes) or less (any vote) common and assume that a replay moves
every visitor's voting in the same direction; intervals are the 2.5th percentile of the lower bound and the 97.5th
percentile of the upper bound over 500 bootstrap draws of visitors. The bounds
for each period of the replay probability use the conversations of that period. Among conversations with exactly one
eligible answer, a decisive vote followed 32.2\% of replays and 29.0\% of
fresh answers, and the eligible answer won 20.6\% and 51.7\% of those votes.
Panel (c): appearances of a stored text from the same model to the same prompt, fresh and replayed, in
conversations with a decisive vote; regressions include the other answer's replay indicator and fixed effects for the
text. The equally weighted row averages, over the 265 texts with decisive votes in both forms,
the difference between the text's win shares when replayed and when fresh; its standard error treats texts as
independent. The thumbs rows use every appearance and add fixed effects for the other model. The mean is the share of
decisive votes won when shown fresh. Standard errors clustered by visitor unless stated.}
\end{table}

%% file: generated/tab_s1_judgments.tex
\begin{table}[t]
\centering
\caption{Study 1: what voters say about a replayed answer (percentage points).}
\label{tab:s1judg}
\small
\begin{tabular}{lccc}
\toprule
 & Mean (\%) & Replayed minus fresh & Same stored text \\
\midrule
\multicolumn{4}{l}{\emph{Qualifier tags attached to the vote (1{,}005 decisive votes)}} \\
\quad Useful & 18.0 & $-$6.3 (4.8) & $-$9.1 (3.1) \\
\quad Complete & 19.5 & $-$21.0 (5.2) & $-$23.2 (3.5) \\
\quad Clear formatting & 16.3 & $-$7.2 (4.0) &  \\
\quad Creative & 5.0 & 1.9 (2.0) &  \\
\quad Superficial & 9.8 & 12.7 (4.0) & 12.2 (3.0) \\
\quad Instructions not followed & 6.5 & 5.5 (3.1) &  \\
\quad Incorrect & 5.8 & 8.0 (2.9) & 7.0 (2.1) \\
\midrule
\multicolumn{4}{l}{\emph{Thumbs on the answer itself (11{,}699 eligible answers)}} \\
\quad Any thumbs reaction & 3.50 & $-$0.21 (0.30) & \\
\quad Thumbs up & 2.17 & $-$1.36 (0.28) & \\
\quad Thumbs down & 1.32 & 1.14 (0.23) & \\
\quad Thumbs-up share among reacted answers & 62.1 & $-$41.4 (4.5) & \\
\quad \quad with model$\times$prompt FE & & $-$32.2 (7.0) & \\
\bottomrule
\end{tabular}
\par\smallskip
\parbox{0.95\linewidth}{\footnotesize Notes: Tag rows use the vote sample and specification of column (3) in
Table~\ref{tab:s1votes}; the dependent variable is the left-minus-right difference in the tag. The last column uses the 1{,}336 decisive
votes on stored texts that appear both fresh and replayed, with fixed effects for the text; the dependent variable is
the difference between the tag on the text and the tag on the other answer. The first three thumbs
rows are answer-level regressions on a replay indicator with week and model$\times$prompt fixed effects and include
every eligible answer. The share of thumbs-up among the 409 answers that drew a
reaction is estimated with week fixed effects only and then with model$\times$prompt fixed effects added. Standard
errors clustered by visitor.}
\end{table}

%% file: generated/tab_s1_robust.tex
\begin{table}[t]
\centering
\caption{Study 1: robustness and heterogeneity of the vote penalty (percentage points).}
\label{tab:s1robust}
\small
\begin{tabular}{lcc}
\toprule
Sample & Replayed minus fresh & Decisive votes \\
\midrule
Main specification (column 3) & $-$25.8 (4.4) & 1{,}005 \\
\midrule
Excluding the prompt with the most votes & $-$24.3 (5.0) & 943 \\
Excluding time-sensitive prompts & $-$24.7 (5.8) & 789 \\
Prompts asked by three or more visitors & $-$24.3 (4.5) & 797 \\
Excluding the shared cookie-less identifier & $-$26.0 (4.8) & 903 \\
Replays that match an earlier answer word for word & $-$27.9 (4.5) & 983 \\
Eligible answers whose key was used within the previous 24 hours & $-$24.0 (4.6) & 905 \\
Eligible answers whose store served a replay within the previous 48 hours & $-$25.4 (10.1) & 202 \\
16 March--5 April (replay share about 0.4) & $-$24.1 (5.8) & 556 \\
6 April--18 May (replay share about 0.7) & $-$27.1 (8.1) & 449 \\
\midrule
Short answers (mean under 300 characters) & $-$19.1 (10.0) & 188 \\
Medium answers (300--1,500) & $-$33.8 (13.5) & 272 \\
Long answers (over 1,500) & $-$48.1 (6.4) & 545 \\
A reasoning model in the pair & $-$19.4 (11.5) & 479 \\
No reasoning model in the pair & $-$28.8 (5.3) & 526 \\
\bottomrule
\end{tabular}
\par\smallskip
\parbox{0.95\linewidth}{\footnotesize Notes: Specification of column (3) in Table~\ref{tab:s1votes}. Standard
errors clustered by visitor in parentheses. Answer length is the mean length of the two first answers. A pooled
interaction of the replay with the expected length of the eligible answer (the mean log length of the earlier fresh
answers from the same model to the same prompt) gives \SoneLenPriorB points per log unit (\SoneLenPriorSE) in this
specification and \SoneLenPriorNophB (\SoneLenPriorNophSE) without prompt fixed effects. In the two stricter
eligibility rows, a conversation is kept only if each of its eligible answers meets the stricter rule.}
\end{table}

%% file: sections/05_study2.tex
\section{Study 2: Answers That Take Longer to Write}
\label{sec:study2}

Study 1 compares answers that appear at once with answers that unfold on screen. Study 2 asks what happens when an
answer that unfolds on screen takes longer than usual. It uses the first exchanges from \StwoStart to \StwoEnd in which both
answers were generated fresh and timed, after the filters in \cref{sec:setting}: \StwoNConv conversations from
\StwoNVisitors identified visitors, involving \StwoNModels models. The median answer took \StwoDurMedian seconds
(interquartile range \StwoDurPTwoFive to \StwoDurPSevenFive seconds), and the median visitor waited \StwoWaitMedian
seconds for the slower of the two answers.

\subsection{Design}

Generation time depends on the model, on how long the answer is and on how much the model reasons before answering.
We measure \emph{slowness} as the part of an answer's log generation time that these do not explain: the residual from
a regression of log duration on log answer length and log reasoning length, estimated separately for each model in
each month. An answer with slowness 0 took as long as is typical for that model and that much text; slowness of 0.7
means it took about twice as long. The standard deviation of slowness is \StwoSlowSd log units.

Slowness can still reflect content, for example if some kinds of text are slower to produce. We therefore use a
measure of congestion as an instrument: the average slowness of the same model's answers to other visitors on the same
day, leaving out the focal visitor's own answers. Because the platform's server and each model provider were shared
by all visitors, a busy day for a provider slowed every answer from that model (\cref{sec:setting}). The regressions
absorb a fixed effect for each calendar day, so the identifying variation is the difference between a model's
congestion and that of other models on the same day; it has a standard deviation of \StwoInstrSdResid log units after
the fixed effects. Congestion is measured net of answer length because raw generation times also carry the day's mix
of requests: a congestion measure built from raw times predicts longer answers from the focal models
(\StwoRawBalLcmLogB log units per log unit, standard error \StwoRawBalLcmLogSE), whereas the length-adjusted measure does not
(\cref{tab:ivbal}). We report ordinary least squares (OLS) estimates, which use all variation in slowness, and
instrumental-variable (IV) estimates, which use only the variation driven by congestion. The IV estimates rest on two
assumptions. The first is that, given the
fixed effects, a model's congestion on a given day is unrelated to who visited and what they asked. The second is that
congestion affects what visitors do only through the time the answer takes.

The congestion measure strongly predicts an answer's own slowness (a coefficient of \StwoFsCoef, first-stage $F$
statistic \StwoContIvF), and it is unrelated to the characteristics of the conversation that it should not affect
(\cref{tab:ivbal} in the appendix). On congested days prompts were no longer, visitors were no more experienced, and
topics and modes did not shift. Answers were no shorter, but they contained less reasoning text (\cref{tab:ivbal}). Because
congestion could change the answer itself in this way, we also report estimates without the length and reasoning
controls in the regression and with the log generation time itself as the regressor (\cref{tab:s2robust}).

For continuation, the unit is the conversation and the regressor is the mean slowness of the two answers. All
regressions absorb fixed effects for the pair of models in each month, the calendar day, the hour of the week and the
topic of the conversation, and they control for answer length, reasoning length, prompt length and the visitor's
experience on the platform. Within-visitor estimates add a fixed effect for each visitor, so that they compare the
same person's faster and slower conversations. For votes, the regressor is the difference in slowness between the
left and right answers, and the instrument is the difference in the two models' congestion. The main vote
specification absorbs a fixed effect for the ordered pair of models in each month, which accounts for which model sat
on the left, and a calendar-day fixed effect. Because many visitors share the same day's congestion, standard errors
are clustered two ways, by visitor and by calendar day \citep{cameron2011robust}.

\subsection{Waiting earns no detectable credit}

\Cref{tab:s2votes} reports the vote regressions. Among two answers unfolding on screen, the one that took longer for
its length did not win more often. In the main specification, one log unit of slowness changed the chance of winning
by \StwoVoteOlsOrderedB points (\StwoVoteOlsOrderedSE), and by \StwoVoteIvOrderedB points (\StwoVoteIvOrderedSE) when
slowness is driven by provider congestion. Neither estimate is distinguishable from zero, and for a doubling of the
wait
the 95\% interval of the IV estimate runs from \StwoVoteIvOrderedDblLo to \StwoVoteIvOrderedDblHi points, which excludes large gains. A specification
with only the unordered pair of models gives negative estimates (OLS \StwoVoteOlsB, \StwoVoteOlsSE; IV \StwoVoteIvB,
\StwoVoteIvSE). The difference comes from the ordering of the pair, which accounts for which model sat on the left,
and not from the calendar-day effects: with ordered pairs alone, the estimates are \StwoVoteOlsOpmB and
\StwoVoteIvOpmB points. For comparison, an answer that is one log unit longer in text wins \StwoVoteOlsDlcAbs points
more often, the familiar length effect in preference data \citep{singhal2024long}, and a replay in Study 1 lost
\SoneVoteVThreeAbs points.

Reactions to individual answers point the same way. Holding the slowness of the answer beside it fixed, a slower
answer drew fewer thumbs-up (\StwoRxpLikedOwnBx points per log unit, \StwoRxpLikedOwnSEx) and more thumbs-down
(\StwoRxpDislikedOwnBx, \StwoRxpDislikedOwnSEx), from bases of \StwoRxpLikedOwnMean\% and \StwoRxpDislikedOwnMean\%.
The answer beside a slow one drew more thumbs-up (\StwoRxpLikedPartnerBx, \StwoRxpLikedPartnerSEx): while the visitor
waited for the slower answer, they could already react to the one that had arrived.

The vote thus gives no detectable support to either H2a or H2b: waiting longer for an answer that is already being
written did not make it win more often. This is the pattern progressive display predicts, and the interval above bounds
the net effect of the two readings of time (\cref{tab:predictions}). On individual answers, a slower
answer drew slightly more thumbs-down, as a cost of waiting would imply. The laboratory evidence varies a
different part of the wait. \citet{tan2026latency} delayed the first word and then streamed the answer at a fixed pace;
a delay of 9 rather than 2 seconds raised perceived thoughtfulness, and 20 seconds added nothing beyond 9. Our measure
covers the whole time to the last word of an answer that took tens of seconds to write, so the two sets of estimates
describe different timing changes and do not lie on one estimated curve.

\input{generated/tab_s2_votes}

\subsection{Waiting costs continuation}

Waiting does matter for what visitors do next. \Cref{tab:s2engage} shows that a slower first exchange lowered the
chance that the visitor asked a second question by \StwoContOlsAbs points per log unit of slowness (\StwoContOlsSE),
from a base of \StwoContMean\%, or about \StwoContRelOls\% of the base rate. The congestion-based estimate is similar
(\StwoContIvAbs points, \StwoContIvSE). Comparing the same visitor's conversations, the estimates are
\StwoContOlsVisAbs (\StwoContOlsVisSE) and \StwoContIvVisAbs points (\StwoContIvVisSE). A doubling of the wait thus
lowers the chance of a follow-up question by about \StwoContDoubleOls points. \Cref{tab:s2robust} in the appendix shows
that the point estimate is the same without the length and reasoning controls in the regression and with the log
generation time rather than slowness as the regressor (\StwoContIvRawexpB, \StwoContIvRawexpSE). It is also the same with
an instrument that leaves out only the focal conversation, with the wait for the slower answer in log seconds as the
regressor, and in conversations without reasoning models. Clustering by visitor and week instead of visitor and day, to allow
congestion shocks to persist over several days, raises the standard error of the IV estimate to \StwoContIvClwkSE (OLS
\StwoContOlsClwkSE). The reduced form, the effect of the congestion measure itself, is \StwoContRfB points per log unit
(\StwoContRfSE). Over the variation the instrument actually provides, one standard deviation of congestion lengthens
the time the answers take by about \StwoInstrSdExpPct\% and changes the chance of a follow-up question by \StwoInstrSdCont points.

\Cref{fig:dose}a shows how continuation varies across the range of slowness. After adjusting for the same fixed
effects and controls (OLS), the share of
visitors who asked a second question in each of the fastest eight tenths of conversations, whose median waits run from
\StwoBinZeroWait to \StwoBinSevenWait seconds, differs from the fastest tenth by less than half a point. It is
\StwoDecEightAbs points lower in the ninth tenth (\StwoDecEightSE; median wait \StwoBinEightWait seconds) and
\StwoDecNineAbs points lower in the slowest tenth (\StwoDecNineSE; median wait \StwoBinNineWait seconds). A regression
with separate slopes below and above the 80th percentile of slowness gives \StwoSplineLoB points per log unit
(\StwoSplineLoSE) below and \StwoSplineHiB (\StwoSplineHiSE) above. In these adjusted comparisons, moderately slow answers go with
little loss, and the loss is concentrated in the long waits of the slowest fifth of conversations.

The cost recurs later in a conversation. Among visitors who had already continued, a slower answer at turns 2 to 6
lowered the chance of yet another question by \StwoContkOlsAbs points per log unit (\StwoContkOlsSE), and by
\StwoContkOlsVisAbs points within visitor (\StwoContkOlsVisSE), from a base of \StwoContKMean\%. The congestion
estimate is \StwoContkIvB points (\StwoContkIvSE).

Other outcomes move differently. Slower answers did not change the chance of a final vote (IV \StwoVotedIvB,
\StwoVotedIvSE), and the estimates for returning to the platform on another day within a week are small and not
distinguishable from zero once congestion is used (\StwoReturnSevendIvCensB, \StwoReturnSevendIvCensSE). Slower
answers raised the chance that the visitor reacted to an answer (IV \StwoReactedIvB, \StwoReactedIvSE), which matches
the reactions to the waiting partner described above. These results support H3: the cost of waiting shows up where the user decides
whether to ask again.

\input{generated/tab_s2_engage}

\begin{figure}[t]
\centering
\includegraphics[width=\linewidth]{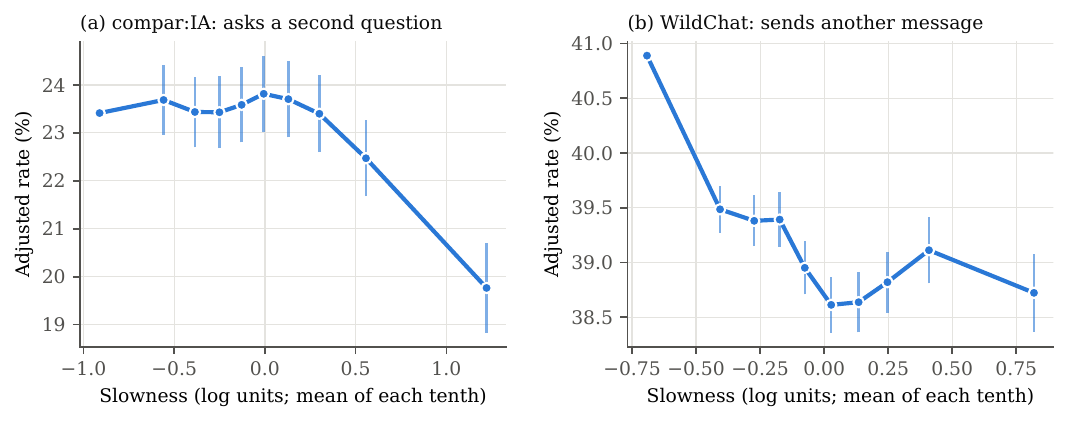}
\caption{Continuation by tenth of slowness. Panel (a): compar:IA first exchanges; the rate at which visitors asked a
second question, adjusted for the fixed effects and controls of \cref{tab:s2engage}. Panel (b): WildChat replies; the
rate at which users sent another message, adjusted for the fixed effects of \cref{tab:s3}. Bars are 95\% intervals for
each tenth's difference from the fastest tenth, with standard errors clustered as in the main tables. Slowness in
panel (a) is log time relative to similar answers from the same model, so 0.7 means about twice as long as usual; in
panel (b) it is measured on the log of one plus the latency in seconds.}
\label{fig:dose}
\end{figure}

%% file: generated/tab_s2_votes.tex
\begin{table}[t]
\centering
\caption{Study 2: waiting earns no detectable credit in the vote (percentage points per log unit of slowness).}
\label{tab:s2votes}
\small
\begin{tabular}{lcccc}
\toprule
 & OLS & IV & First-stage F & Observations \\
\midrule
\multicolumn{5}{l}{\emph{Blind vote: the left-hand answer wins (slowness, left minus right)}} \\
Model pair $\times$ month FE, ordered, and calendar-day FE & 0.2 (0.4) & $-$2.9 (2.5) & 469 & 80{,}205 \\
\quad standard errors clustered by visitor and week &  & $-$2.9 (2.8) & 232 & 80{,}205 \\
\quad without calendar-day FE & 0.2 (0.4) & $-$2.7 (2.5) & 466 & 80{,}205 \\
Model pair $\times$ month FE, unordered & $-$1.0 (0.4) & $-$6.6 (3.2) & 159 & 80{,}205 \\
\quad adding calendar-day FE & $-$1.0 (0.4) & $-$6.5 (3.2) & 159 & 80{,}205 \\
\quad without length and reasoning controls &  & $-$7.4 (3.5) & 152 & 80{,}205 \\
\quad one-turn conversations & $-$0.7 (0.5) & $-$4.9 (3.3) & 155 & 66{,}321 \\
\midrule
\multicolumn{5}{l}{\emph{Thumbs on each streamed first answer}} \\
Thumbs up on the answer & $-$0.04 (0.06) & 0.47 (0.36) & 1{,}609 & 659{,}969 \\
Thumbs down on the answer & 0.20 (0.04) & 0.61 (0.21) & 1{,}609 & 659{,}969 \\
Thumbs-up share among reacted answers & $-$1.6 (0.4) & & & 57{,}299 \\
\bottomrule
\end{tabular}
\par\smallskip
\parbox{0.95\linewidth}{\footnotesize Notes: Vote regressions control for the left-minus-right difference in log
length and log reasoning length unless stated. The ordered specification gives each arrangement of two models on
the left and right its own fixed effect; in the unordered specification, one log unit of extra length wins
\StwoVoteOlsDlcB points (\StwoVoteOlsDlcSE). The instrument is the left-minus-right difference in the two models'
same-day slowness among other visitors. Answer-level regressions absorb model $\times$ month and calendar-day fixed
effects and control for length and reasoning length; they include every streamed first answer with a recorded time,
including answers whose partner was not timed. Standard errors clustered two ways, by visitor and by calendar day.}
\end{table}

%% file: generated/tab_s2_engage.tex
\begin{table}[t]
\centering
\caption{Study 2: slower first answers lead to fewer follow-up questions (percentage points per log unit of slowness).}
\label{tab:s2engage}
\small
\begin{tabular}{lcccc}
\toprule
 & OLS & IV & OLS & IV \\
 & \multicolumn{2}{c}{All conversations} & \multicolumn{2}{c}{Within visitor} \\
\midrule
Asks a second question & $-$1.8 (0.2) & $-$2.0 (1.0) & $-$1.5 (0.2) & $-$2.4 (1.1) \\
Returns on another day within 7 days & $-$0.1 (0.1) & $-$0.6 (0.5) &  &  \\
Casts a final vote & 1.3 (0.3) & 0.0 (0.8) &  &  \\
Reacts to a message & 0.3 (0.1) & 1.7 (0.7) &  &  \\
\midrule
Mean of ``asks a second question'' (\%) & \multicolumn{4}{c}{\StwoContMean} \\
First-stage coefficient (F statistic) & & \StwoFsCoef\ (\StwoContIvF) & & (\StwoContIvVisF) \\
Conversations & \StwoContOlsN & \StwoContIvN & \StwoContOlsVisN & \StwoContIvVisN \\
\midrule
\multicolumn{5}{l}{\emph{Later turns: continues after turn 2--6 (mean \StwoContKMean\%)}} \\
Continues the conversation & $-$3.7 (0.3) & $-$1.6 (1.5) & $-$4.2 (0.4) & \\
Turns & \StwoContkOlsN & \StwoContkIvN & \StwoContkOlsVisN & \\
\bottomrule
\end{tabular}
\par\smallskip
\parbox{0.95\linewidth}{\footnotesize Notes: Slowness is the mean of the two answers' log generation time net of
their length and reasoning length within model and month. The instrument is the mean slowness of the same models'
answers to other visitors on the same day (provider congestion). All columns absorb model pair $\times$ month,
calendar day, hour of week and topic fixed effects and control for answer length, reasoning length, prompt length and
visitor experience; within-visitor columns add visitor fixed effects. Seven-day returns use the 319{,}912
conversations with seven days of follow-up in the release. Standard errors clustered two ways, by visitor (by session for visitors without a
cookie) and by calendar day.}
\end{table}

%% file: sections/06_study3.tex
\section{Study 3: Waiting in Everyday Chatbot Use}
\label{sec:study3}

compar:IA is a comparison site, and its visitors know they are testing models. Study 3 asks whether the cost of
waiting appears in ordinary use of a single chatbot. WildChat users asked ChatGPT for help with whatever they
wanted, one model at a time, with no ratings and no reveal \citep{zhao2024wildchat}.

\subsection{Design}

For each reply we compute the latency as the time WildChat's server received the full reply minus the time the
provider created the completion, both recorded to the second (\cref{sec:setting}). Of the \SthreeAuditN replies in the
release, none has a negative latency, \SthreeAuditZero have a latency below one second, recorded as zero, and only
\SthreeAuditOverSixZeroZero exceed ten minutes. Latency tracks reply length closely: within model and month, one log
unit of additional reply length goes with \SthreeElasticity log units of additional latency, as expected if latency
covers the writing of the complete reply. We keep replies with latencies of up to ten minutes and remove hashed IP
addresses that created more than 50 conversations on a single day. The sample contains \SthreeNReplies replies in
\SthreeNConversations conversations from \SthreeNIps hashed IP addresses, from \SthreeStart to \SthreeEnd. The median
reply took \SthreeLatMedian seconds (interquartile range \SthreeLatPTwoFive to \SthreeLatPSevenFive seconds; 95th
percentile \SthreeLatPNineFive seconds). The outcome is whether the user sent another message in the same
conversation after the reply. Users did so after \SthreeContMeanFirst\% of first replies and after
\SthreeContMeanLater\% of later replies.

The design mirrors Study 2 with one change. Short replies carry a fixed overhead and many latencies are recorded as
zero, so we measure slowness on the log of one plus the latency in seconds and adjust for content non-parametrically.
Slowness is that log minus its average among replies from the same model in the same month with similar reply
length, similar message length and the same position in the conversation. Congestion is the average slowness of the
same model's replies to other users in the same clock hour. OpenAI's service was shared by all WildChat users and by
the provider's other customers, so hourly congestion varies for reasons outside any one user's control.

Congestion also tracks demand: with calendar-day and hour-of-week effects, congested hours carried shorter messages
and replies and fewer users in the United States (\cref{tab:ivbal}). The preferred specification therefore absorbs a
fixed effect for each of the \SthreeNHours clock hours. These effects absorb demand shared by all models in an hour, so
the comparison is between models used in the same hour; demand specific to one model remains. The
\SthreeHcShareMultimodel\% of replies sent in hours when more than one model was in use provide the identifying
variation. After these fixed effects, the congestion measure has a standard deviation of \SthreeHcInstrSdResidHour
log units and still strongly predicts a reply's own slowness (coefficient \SthreeHcFsHourCoef, first-stage $F$
statistic \SthreeHcFsHourF). The differences in message and reply length disappear, and the coefficient for the share of
users in the United States shrinks from \SthreeIvbalUsB to \SthreeBalhourUsB (\SthreeBalhourUsSE). We also report estimates that compare
replies sent from the same hashed IP address, which removes stable differences between users, and estimates with
calendar-day and hour-of-week effects in place of the clock-hour effects. All regressions absorb fixed effects for
model by month, position in the conversation and the length cells, and standard errors are clustered two ways, by
hashed IP address and by clock hour.

\subsection{Results}

\Cref{tab:s3} shows that slower replies made users less likely to continue. With clock-hour effects, one unit of
slowness lowered the chance of another message by \SthreeContOlsHourAbs points (\SthreeContOlsHourSE), and by
\SthreeContIvHourAbs points (\SthreeContIvHourSE) when slowness is driven by congestion, from a base of
\SthreeContMean\%. Comparing replies from the same hashed IP address in the same design gives
\SthreeHcContOlsHourIpAbs (\SthreeHcContOlsHourIpSE) and \SthreeHcContIvHourIpAbs points (\SthreeHcContIvHourIpSE).
With calendar-day and hour-of-week effects instead, the estimates are \SthreeContOlsAbs and \SthreeContIvAbs points
(\SthreeContOlsIpAbs and \SthreeContIvIpAbs within hashed IP address). Within the clock-hour design, the congestion
estimate comes mainly from later replies (\SthreeHcContIvLaterHourB, \SthreeHcContIvLaterHourSE) and from GPT-3.5 models
(\SthreeHcContIvGptThreeFiveHourB, \SthreeHcContIvGptThreeFiveHourSE). For first replies, GPT-4-class models and users in the United States, the clock-hour IV estimates are
\SthreeHcContIvFirstHourB (\SthreeHcContIvFirstHourSE), \SthreeHcContIvGptFourHourB (\SthreeHcContIvGptFourHourSE) and
\SthreeHcContIvUsHourB points (\SthreeHcContIvUsHourSE), and the OLS estimates for first replies and GPT-4-class models
are negative. The estimate is similar without the replies recorded as zero
seconds (\SthreeHcContIvHourNozeroB, \SthreeHcContIvHourNozeroSE), which make up \SthreeHcZeroShare\% of the sample and
all fall in the fastest four tenths of slowness. Among users who did continue, a slower reply lengthened the time from
its receipt to the creation of their next request (\SthreeHcGapOlsHourB log points per unit of slowness,
\SthreeHcGapOlsHourSE). \Cref{fig:dose}b shows the adjusted continuation rate by tenth of slowness. Unlike in
compar:IA, most of the fall comes early: users were \SthreeDecOneAbs points (\SthreeDecOneSE) less likely to continue
after a reply in the second-fastest tenth than after one in the fastest tenth, and \SthreeDecNineAbs points
(\SthreeDecNineSE) less likely after one in the slowest tenth.

\input{generated/tab_s3}

Study 3 supports H3 outside the arena. In everyday use of a single chatbot, as on a comparison site, a longer wait for
an answer makes people less likely to keep the conversation going.

%% file: generated/tab_s3.tex
\begin{table}[t]
\centering
\caption{Study 3: in everyday ChatGPT use, slower replies end conversations (percentage points per unit of slowness).}
\label{tab:s3}
\small
\begin{tabular}{lcccc}
\toprule
 & (1) OLS & (2) IV & (3) OLS & (4) IV \\
 & & & \multicolumn{2}{c}{Within hashed IP address} \\
\midrule
\multicolumn{5}{l}{\emph{Exact clock-hour fixed effects (preferred)}} \\
\quad User sends another message & $-$1.3 (0.1) & $-$1.8 (0.7) & $-$2.4 (0.1) & $-$6.7 (1.2) \\
\quad First-stage F &  & 10{,}068 &  & 3{,}950 \\
\multicolumn{5}{l}{\emph{Calendar-day and hour-of-week fixed effects}} \\
\quad User sends another message & $-$1.5 (0.1) & $-$3.5 (0.4) & $-$2.6 (0.1) & $-$6.1 (0.7) \\
\quad First-stage F &  & 58{,}073 &  & 12{,}754 \\
\midrule
Mean (\%) & \multicolumn{4}{c}{\SthreeContMean} \\
Replies & \multicolumn{4}{c}{4{,}378{,}514} \\
\bottomrule
\end{tabular}

\medskip
\begin{tabular}{lccc}
\toprule
Subsample, exact clock-hour fixed effects & OLS & IV & Replies \\
\midrule
First reply in a conversation & $-$0.4 (0.1) & $-$1.2 (1.1) & 2{,}687{,}209 \\
Later replies & $-$2.2 (0.1) & $-$3.6 (0.9) & 1{,}691{,}305 \\
GPT-3.5 models & $-$3.0 (0.2) & $-$1.9 (0.2) & 1{,}232{,}870 \\
GPT-4-class and o1 models & $-$0.3 (0.1) & 1.3 (1.4) & 3{,}145{,}644 \\
Users in the United States &  & 1.4 (1.8) & 910{,}395 \\
Replies that took at most five minutes &  & $-$1.8 (0.7) & 4{,}378{,}055 \\
Excluding latencies recorded as zero seconds & $-$1.8 (0.1) & $-$2.1 (0.7) & 4{,}036{,}299 \\
Standard errors clustered by hashed IP only &  & $-$1.8 (0.6) & 4{,}378{,}514 \\
\midrule
Time to the next request (log seconds, continued replies) & 0.170 (0.007) & & 1{,}650{,}194 \\
\bottomrule
\end{tabular}
\par\smallskip
\parbox{0.95\linewidth}{\footnotesize Notes: Slowness is log(1 + latency in seconds) minus its mean among replies from the
same model in the same month with similar reply length (twentiles), user-message length (deciles) and position in the
conversation. The instrument is the mean slowness of the same model's replies to other users in the same clock hour.
All regressions absorb model $\times$ month, turn-position and length-cell fixed effects. The preferred rows add one
fixed effect per clock hour (19{,}077 hours), so that only differences between models used in
the same hour remain; 59\% of replies were sent in hours with
more than one model. The second panel of rows uses calendar-day and hour-of-week fixed effects instead. Standard errors
clustered two ways, by hashed IP address and by clock hour. The time-to-next-request row is in log units of seconds,
not percentage points, and measures the time from the receipt of the reply to the creation of the completion for the
user's next message. 7{,}750 conversations have no first reply in the sample, because
their first reply fell in a length cell with fewer than 20 replies, lacked the congestion measure or exceeded the
latency limit.}
\end{table}

%% file: sections/07_discussion.tex
\section{Discussion}
\label{sec:discussion}

The three studies separate two things that usually travel together: how an AI answer is delivered and how long the
user waits for it. Delivery moved the vote. Weighting each text equally, the same stored text from the same model won
\SoneIdentEqwFresh\% of its decisive votes when it unfolded on screen and \SoneIdentEqwReplay\% when it filled the
screen at once. Voters explained their choices against the replayed text by calling it less complete, more superficial
and more often incorrect. Waiting moved behavior. Among answers that unfolded on screen, extra waiting
brought no detectable net credit in the vote, and long waits made people less likely to keep asking, both on a public
arena and in everyday ChatGPT use.

\subsection{Which account fits the evidence}
\label{sec:fit}

\Cref{tab:accounts} sets each account against the results that bear on it. Votes that judge only the text, and votes
that favor faster answers, both fail the first test: the answer shown at once lost, by a wide margin, and voters did not
describe it as the more reliable one. Progressive display passes it. On the time an unfolding answer takes, neither
reading of time finds detectable support in the vote, which is what progressive display predicts. The cost of time appears instead in whether users
asked again.

Several features of the setting could produce a penalty for answers shown at once without any of these accounts, and
the data rule them out. The penalty appears on single answers when both answers in a conversation were replays, so it
does not require a visibly working answer on the other side of the screen, although the contrast with such an answer
made it larger. Among answers that unfolded, the one that finished later did not win more often, so the penalty is not a
recency effect. The identical text drew it, and it survived the exclusion of prompts that depend on the date, so stored
content does not explain it. Replays did not change whether visitors voted or, across all conversations, how often they
declared a tie. Among those who voted, a single replay made a tie less likely, which sharpens the comparison rather than
selecting who votes.

\begin{table}[t]
\centering
\caption{The accounts against the evidence.}
\label{tab:accounts}
\small
\begin{tabularx}{\linewidth}{>{\raggedright\arraybackslash}p{0.2\linewidth}>{\raggedright\arraybackslash}X>{\raggedright\arraybackslash}X}
\toprule
Account & Prediction & What the studies found \\
\midrule
Votes judge only the text & The same answer does equally well shown at once or unfolding & \SoneIdentEqwFresh\% against \SoneIdentEqwReplay\% of decisive votes for the identical text; \SoneVoteVThreeB points in the lottery \\
Speed as fluency or competence & The answer shown at once does at least as well and seems at least as reliable & It did worse, and voters called it incorrect more often (\SoneTagIncorrectB points) \\
Progressive display (effort or reading) & The answer shown at once does worse; extra time on an unfolding answer adds nothing & Consistent with both \\
\addlinespace
Time as deliberation & An unfolding answer that takes longer wins more & No detectable net gain: for a doubling of the wait, \StwoVoteIvOrderedDblLo to \StwoVoteIvOrderedDblHi points (95\% interval) \\
Time as a cost & An unfolding answer that takes longer wins less; users ask again less often & No detectable net loss in the vote; fewer follow-up questions on compar:IA and in WildChat \\
\midrule
On-screen contrast only & No penalty when both answers appear at once & Thumbs-up \SoneRxCellLikedCOneOneB points for a replay beside another replay \\
The last answer to finish wins & Slower unfolding answers win more & They do not (\StwoVoteOlsOrderedB OLS, \StwoVoteIvOrderedB IV per log unit) \\
Stored content differs & No penalty for the identical text, or a penalty only for prompts that depend on the date & Penalty for the identical text; \SoneRobTimesensB points without time-sensitive prompts \\
Replays change who votes & Voting or ties change across all conversations & Neither changed; Lee bounds exclude zero \\
\bottomrule
\end{tabularx}
\end{table}

\subsection{Implications for theory}

The labor illusion holds that people value a service more when they can see the work behind it, and its experiments
showed that the display, not the amount of actual work, carried the effect \citep{buell2011labor}. Later work found
that how AI text appears on screen changes laboratory ratings of its quality, by a small margin
\citep{zhou2025thoughtful}, and changes users' willingness to accept it \citep{jin2026word}. Our results carry these
findings into the setting where judgments of AI output have consequences: blind choices that rank models, made by
people who chose their own questions, about the same model's verbatim answer delivered under an operator's real
policy. There, delivery moved the outcome by a wide margin, and in the direction opposite to the one the platform's
designers expected \citep{termignon2026comparia}. Research on the discount that consumers apply to AI-made products
traces it in part to beliefs about the love and effort that went into a product an algorithm created
\citep{rix2025algorithm}. Here both answers came from AI and the wording was fixed, so delivery alone moved the outcome. Revealing part of a generated output before
registration raises engagement \citep{jiang2026seeing}; our results add that how the whole answer is revealed changes
how it is judged. The prediction that machines are judged more reliable when fast
\citep{efendic2020slow} did not carry over to text that users could see unfold: the answer shown at once was the one
voters described as incorrect more often. For generative AI, whose answers unfold by default, the practical question the
labor illusion raises is reversed: not whether to add a display of work, but when back-end decisions such as caching
take it away. Which route carries the effect is the theoretical question the field data leave open. The answer decides whether the effect reflects an inference about effort, which telling users that the pace is set by software
should weaken, or an easier reading, which it should not.

The results also bound the reading of time as information about machine output. Among answers that unfolded on
screen, slowdowns caused by congestion bought no detectable net credit in the vote. Laboratory evidence that a later
first word raises perceived thoughtfulness \citep{tan2026latency} sits alongside this result rather than against it.
The designs vary different parts of the wait: the laboratory study delayed the first word and held the pace of writing
fixed, the replays in our data removed the unfolding altogether, and congestion lengthened the whole time to the last
word. How a delay before the first word combines with the unfolding of the text remains open, and a design that varies
the two separately would show it. The labor illusion's own experiments point the same way as our estimate: with the same
work displayed, longer waits added no value \citep{buell2011labor}.

For research on response time in information systems, the findings move the cost of waiting from attitudes to a
behavioral margin. Earlier work measured satisfaction, attitudes and intentions
\citep{shneiderman1984response,palmer2002website,galletta2004delays,nah2004tolerable}, and research on waiting
interfaces asked how displays change the perceived wait \citep{lee2012online,lee2017online}. Conversational systems
offer a direct behavioral measure, whether the user asks again after an answer has arrived, and in our data this
measure moved with the wait while preference votes did not. The decision resembles the choice to start another service
encounter after a costly wait rather than the abandonment of a queue in progress
\citep{batt2015waiting,aksin2013structural}. In the adjusted comparisons on compar:IA, as in queues, most of the loss
came from the longest waits; in WildChat, where replies were much faster, it came mostly from moving out of the fastest
tenth.

\subsection{Implications for design}

Caching is attractive to every operator of a generative AI service because it saves computation and time
\citep{bang2023gptcache}. Our results show that how a stored answer is shown matters for how it is judged. An answer
that fills the screen at once is judged worse than the same answer unfolding on screen, so a cache that saves cost can lower
perceived quality. compar:IA's own design intended to replay stored answers in pieces ``to mimic real streaming
behavior'' \citep{comparia2026code}; the replays in our window ran far faster than intended. The results support
testing replay at a readable pace, and they raise the same question for speculative decoding and other methods that
change the pace at which text appears \citep{leviathan2023speculative}. Designed experiments can then identify the pace that serves users best and show
whether a slower display of a stored answer helps users or only raises their opinion of it.

The results do not support adding delay to make answers seem more thoughtful. Among answers that unfolded on screen,
longer waits earned no detectable net credit in the vote and cost follow-up questions. This matters as reasoning models, which spend more time
before answering \citep{guo2025deepseek,snell2024scaling}, become the default. Their extra time is worth spending when
it buys a better answer; in our data it does not buy a better impression on its own.

\subsection{Implications for evaluation}

Preference leaderboards built from blind votes have become a standard yardstick for generative AI
\citep{chiang2024arena,termignon2026comparia}. What delivery means for them depends on what they are meant to measure.
If a leaderboard measures the quality of the text a model writes, delivery is a confound: a model whose answers
happened to be replayed more often would have lost ground for reasons unrelated to its output. Arenas can remove it by
showing every answer under the same display conditions, or record caching, streaming pace and the display of
reasoning and adjust for them, as they already do for answer length \citep{singhal2024long}. If a leaderboard measures
the service a user experiences, delivery belongs in the score, but then the cost of latency that we measure is missing from it.
The vote showed no detectable reward for slower answers, yet slower answers lowered the chance that the user kept the
conversation going. A preference score cannot register this cost when both answers are slow, because a vote compares
the two: a delay that slows both leaves the winner unchanged, while the user still pays it. Continuation measures such as follow-up rates
belong next to preference scores when systems are compared.

\subsection{Limitations and future work}
\label{sec:limits}

Several features of the setting point to directions for future work. First, a replay changes how fast an answer
appears, how it unfolds and how much time the reader has to follow it, all at once, so the field data identify the
effect of this delivery but not the psychological route behind it. The labor illusion proposes inferred effort, while
unfolding text may also help readers follow an answer, hold attention, or make an answer seem written for the request.
A designed experiment could separate these routes: show the same answer on the same schedule, unfolding at a readable
pace or appearing mostly in a final burst, and tell half the readers that software sets the pace independently of the
work the model did. An effect that weakens under this disclosure, while reading comfort holds, points to inferred
effort; one that survives it points to reading. Asking readers how likely a specific, checkable claim is to be correct,
before any overall judgment and with answers of known accuracy, would show whether delivery changes beliefs about
accuracy or only the reasons voters give, which the optional qualifiers in our data cannot settle. Varying the delay
before the first word separately would connect these results to laboratory evidence on onset delay. Second, eligibility for a replay is reconstructed from a public release that omits some conversations, the release
contains only completed answers, and durations are measured on the platform's server rather than in the visitor's browser. Logs of every request and of what each visitor saw would sharpen all three. Third, the congestion estimates rest on the assumption that a provider's load affects visitors only
through the time their answers take. A blocked server can also slow the interface itself, and congestion slightly
shortened reasoning text. In WildChat, a fixed effect for each clock hour absorbs demand shared by all models in that
hour but not demand specific to one model. Providers' own records of outages and load would offer cleaner variation. Fourth,
continuation is an action, not a measure of satisfaction or benefit: a user may stop because a question was answered.
Linking continuation to task success would show what the lost questions were worth. Fifth, WildChat identifies users
only by hashed IP addresses, and its public release omits conversations that contain toxic content anywhere, which
may relate to how long they ran. Finally, the replay lottery covers prompts that were asked repeatedly, which are
shorter and more often suggested or assigned than the average prompt. compar:IA's audience is largely French-speaking
and includes many students, and WildChat users chose a free service. Replications with unique, demanding questions,
in commercial assistants and in other languages would show how far the results extend.

\section{Conclusion}
\label{sec:conclusion}

Is an AI answer worth the wait? In these settings, users gave no detectable net credit in their votes for extra waiting
on answers they watched being written, and long waits made them less likely to keep asking. They did, however, judge answers that
appeared at once more harshly than the same answers unfolding on screen. What moved their votes was how an answer was
delivered, not how long it took. For those who build and evaluate AI systems,
two tests follow: whether showing stored answers at a readable pace removes the penalty, and whether shorter waits keep
users asking. Preference votes register the first and miss the second.

%% file: sections/08_appendix.tex
\section{Additional Results}
\label{app:tables}

\subsection{Study 1}

\Cref{tab:s1engage} reports how replays relate to continuation and feedback in Study 1. The regressor counts the
replayed first answers in a conversation, and the fixed effects interact the number of eligible answers with the week,
so that the comparison is between conversations whose eligible answers happened to be replayed or not. The last row
shows that a conversation with exactly one replayed answer is less likely to end in a vote declaring the two answers
equal.

\input{generated/tab_s1_engage}

\subsection{Study 2}

\Cref{tab:ivbal} relates the congestion instruments to characteristics of the conversation that congestion should not
affect. For compar:IA, same-day congestion of a model is unrelated to the length of the prompt, the experience of the
visitor, the share of custom model choices and the length of the answer. On congested days answers contained less reasoning text. For WildChat, congested hours carried shorter messages and
replies and fewer users in the United States when the regression holds only the day and hour of the week fixed. With a
fixed effect for each clock hour, the length differences disappear and the difference in the share of users in the
United States shrinks (\cref{sec:study3}). \Cref{tab:s2robust} reports alternative specifications of the follow-up result, and
\cref{tab:s2topics} reports the vote regression separately by topic and for the qualifier tags. The topic estimates are
exploratory and range from \StwoTopicMin to \StwoTopicMax points per log unit across the \StwoTopicN topics; only one positive estimate, for culture and arts (\StwoTopicCultureB, \StwoTopicCultureSE), is about two
standard errors from zero, which is not unusual among ten estimates without adjustment for multiple comparisons.

\input{generated/tab_ivbal}
\input{generated/tab_s2_robust}
\input{generated/tab_s2_topics}

\section{Implementation of the Cache and the Server}
\label{app:implementation}

\Cref{tab:impl} lists the changes to the platform's public source code \citep{comparia2026code} that define the two
designs, with links to the pull requests and commits. The repository records when code was written and merged, not
when it was deployed. The merge dates are therefore inferred deployment dates; for the cache, the data confirm the date,
since the first replay appears on 16 March 2026, the day the cache was merged.

\begin{table}[t]
\centering
\caption{compar:IA code relevant to the designs, from the public repository.}
\label{tab:impl}
\small
\begin{tabularx}{\linewidth}{lX}
\toprule
Merged & Change \\
\midrule
16 March 2026 & Cache of first-turn answers. Key: model and opening prompt, with surrounding spaces removed and
capital letters ignored. Up to five answers per key, expiring 48 hours after the key was last written or read. On each
first-turn request, a uniform random draw decides whether to look for the key; if found, one stored answer is chosen
at random and replayed. The probability is a deployment setting (0.5 by default in the current code). Pull request
\href{https://github.com/betagouv/ComparIA/pull/379}{\#379}, commit \href{https://github.com/betagouv/ComparIA/commit/83eb46142ac0}{83eb461}. \\
16 March 2026 & Replay display: the stored text is sent in pieces of at least 20 characters (about fifteen pieces
for longer answers) with a pause of 0.2 seconds after each; stored reasoning text is sent at once. If a pause is
cancelled, the replay continues with the next piece at once. The recorded duration runs from the start of the replay
to its last piece. Same pull request. \\
16 March 2026 & Answers are stored from the final text of the answer as shown. Commit
\href{https://github.com/betagouv/ComparIA/commit/7d5db5019d6a}{7d5db50}. \\
12 October 2026 & Fix for a server that ``held the one event loop'' while waiting for a provider, so that a slow
model delayed every visitor's requests. A second change lets each answer stream at its own pace; its description
notes that until then ``the slower read was cancelled as soon as the other model answered,'' and that the cached
replay had ignored cancellation ``only to survive those cancelled reads.'' Pull request
\href{https://github.com/betagouv/ComparIA/pull/795}{\#795}, commits \href{https://github.com/betagouv/ComparIA/commit/2364b6c8cdb0}{2364b6c} and
\href{https://github.com/betagouv/ComparIA/commit/34ae6bc7f052}{34ae6bc}. \\
\bottomrule
\end{tabularx}
\end{table}

The replay share in the data, \SoneShareCachedRegimeOne\% of eligible answers until 5 April and
\SoneShareCachedRegimeTwo\% afterwards, is consistent with a deployed probability of about 0.4 and then about 0.75.
The second change on 12 October explains why the replays in our window were fast. On every round of the reading loop,
the server read the two answers of a comparison together and cancelled the slower read once the other answer arrived;
the replay's pause was such a read, so it ended whenever the other answer produced text.

\section{Reconstructing Replay Eligibility}
\label{app:eligibility}

We reconstruct, for every first answer from the first recorded replay onward, whether the store held answers for its
key. An answer is eligible if two conditions hold. First, the same model gave at least one fresh first answer to the same
opening prompt after caching began and before this answer, with prompts compared as the platform compared them, after
removing surrounding spaces and ignoring capital letters. Second, the previous first answer for that model and prompt,
fresh or replayed, came no more than 48 hours earlier. Two sources of error remain. Conversations that are not in the public release, for
example because they were removed for containing personal data \citep{termignon2026comparia}, may have created or
refreshed a key that we do not observe. Some answers we classify as ineligible were therefore eligible:
\SoneAuditCachedIneligNorm of the \SoneAuditNCached replays in the window (\SoneAuditIneligShare\%) fall in this group
and are not used. Conversely, a key may have expired early or failed to be written, so that some answers we classify
as eligible could not have been replayed; these are always fresh, and comparing answers to the same prompt limits the
damage they can do. Restricting the vote sample to answers whose key had been used within the previous 24 hours, or to
answers whose store had served a replay within the previous 48 hours, leaves the estimate about the same
(\cref{tab:s1robust}). Matching prompts by their exact text rather than by the platform's rule identifies
\SoneAuditNEligExact rather than \SoneAuditNEligNorm eligible answers across all traffic. Of the \SoneNEligCells
model--prompt cells in our sample, \SoneCellsMixed contain both replayed and fresh answers; they hold
\SoneAnswersInMixedCells of the \SoneNEligAnswers eligible answers and provide the comparisons within prompt.

\section{Sample Construction}
\label{app:flow}

\Cref{tab:flow} traces each analysis sample from the public releases. For compar:IA, every study starts from the
conversations in the five standard modes without automated traffic. Study 1 then keeps the conversations in the replay
window whose two first answers were both recorded, and its vote and answer-level samples follow from the eligibility
rule of \cref{app:eligibility}. Study 2 keeps first exchanges in which both answers were generated fresh, timed and
long enough to adjust for length, and in which the congestion measure could be computed. For WildChat, the filters
remove replies with implausible latencies or empty text and heavy automated users, and then require enough similar
replies to adjust for content and enough replies from other users in the same hour to measure congestion.

\begin{table}[t]
\centering
\caption{Analysis samples.}
\label{tab:flow}
\small
\begin{tabular}{lr}
\toprule
\multicolumn{2}{l}{\emph{compar:IA (release to 18 May 2026)}} \\
Conversations in the release & \SampNConvAll \\
\quad in the five standard modes & \SampNStdMode \\
\quad after removing the \SampNAutoConvStd conversations of \SampNAutoIds automated identifiers & \SampNStdNonauto \\
\multicolumn{2}{l}{\emph{Study 1: \SoneStart to \SoneEnd}} \\
\quad conversations with both first answers recorded & \SoneNConvEra \\
\quad with at least one eligible first answer & \SoneNConvElig \\
\quad with both first answers eligible & \SoneNConvBothElig \\
\quad final votes in conversations with an eligible answer & \SoneNVotesEligAll \\
\quad decisive votes (main vote sample) & \SoneNVotesDecisive \\
\quad eligible answers (answer-level analyses) & \SoneNEligAnswers \\
\multicolumn{2}{l}{\emph{Study 2: \StwoStart to \StwoEnd}} \\
\quad conversations with both first answers timed & \SampNWithTiming \\
\quad both fresh, timed and adjustable for length, with the instrument (main sample) & \StwoNConv \\
\quad decisive votes among them & \StwoNVotes \\
\quad later turns (turns 2--6) & \StwoNTurnsLater \\
\quad streamed first answers with a recorded time (answer-level analyses) & \StwoNRxAnswers \\
\midrule
\multicolumn{2}{l}{\emph{WildChat-4.8M (April 2023 to July 2025)}} \\
Replies in the release & \SthreeAuditN \\
\quad latency at most ten minutes, non-empty, from IP addresses with at most 50 conversations a day & \SthreeAuditNPrepared \\
\quad in a content cell of at least 20 replies, with the congestion measure (main sample) & \SthreeNReplies \\
\quad conversations & \SthreeNConversations \\
\quad hashed IP addresses & \SthreeNIps \\
\bottomrule
\end{tabular}
\par\smallskip
\parbox{0.95\linewidth}{\footnotesize Notes: The automated identifiers account for \SampNAutoConv conversations in
all modes, \SampNAutoConvStd of them in the standard modes. The Study 2 timing count includes all modes and traffic; the
main sample applies the filters above it. Answer-level analyses in Study 2 include answers whose partner was not timed, so they
contain more answers than twice the conversation sample.}
\end{table}

%% file: generated/tab_s1_engage.tex
\begin{table}[t]
\centering
\caption{Study 1: replays and engagement (percentage points per replayed answer).}
\label{tab:s1engage}
\small
\begin{tabular}{lccc}
\toprule
 & Mean (\%) & Eligibility $\times$ week FE & + prompt FE \\
\midrule
Asks a second question & 8.8 & 0.8 (0.6) & 0.2 (0.4) \\
Casts a final vote & 22.8 & $-$1.7 (0.7) & $-$0.3 (0.6) \\
Reacts to a message & 5.1 & $-$0.8 (0.4) & 0.2 (0.4) \\
Returns on another day within 7 days & 1.6 & $-$0.5 (0.2) & $-$0.2 (0.2) \\
Votes the two answers equal (among votes) & 36.1 & $-$9.8 (2.6) & \\
\bottomrule
\end{tabular}
\par\smallskip
\parbox{0.95\linewidth}{\footnotesize Notes: 6{,}898 conversations with at least one eligible
first answer. The regressor counts replayed first answers (0, 1 or 2); the fixed effects interact the number of eligible
answers with the week. The tie row uses 1{,}574 votes and an indicator that exactly one
answer was replayed. Standard errors clustered by visitor.}
\end{table}

%% file: generated/tab_ivbal.tex
\begin{table}[t]
\centering
\caption{Congestion instruments and characteristics of the conversation.}
\label{tab:ivbal}
\small
\begin{tabular}{lccc}
\toprule
 & Mean & \multicolumn{2}{c}{Coefficient on the instrument} \\
\cmidrule(lr){3-4}
 & & Calendar-day effects & Exact clock hour \\
\midrule
\multicolumn{4}{l}{\emph{Study 2 (compar:IA): same model, same day}} \\
Prompt length (log characters) & 4.790 & $-$0.045 (0.026) & \\
Visitor's earlier conversations (log) & 0.911 & $-$0.007 (0.026) & \\
Science and technology topic (share) & 0.209 & $-$0.013 (0.008) & \\
Visitor chose the models (share) & 0.261 & 0.006 (0.010) & \\
Answer length (log characters) & 7.467 & $-$0.018 (0.019) & \\
Reasoning length (log characters) & 0.919 & $-$0.091 (0.028) & \\
\midrule
\multicolumn{4}{l}{\emph{Study 3 (WildChat): same model, same hour}} \\
User message length (log characters) & 5.682 & $-$0.276 (0.047) & $-$0.017 (0.049) \\
Reply length (log characters) & 6.346 & $-$0.243 (0.054) & $-$0.047 (0.038) \\
User in the United States (share) & 0.208 & $-$0.051 (0.008) & $-$0.023 (0.010) \\
\bottomrule
\end{tabular}
\par\smallskip
\parbox{0.95\linewidth}{\footnotesize Notes: Each row regresses the characteristic on the congestion instrument
with calendar-day fixed effects (Study 2, the main specification: model pair $\times$ month, day, hour of week; Study 3:
model $\times$ month, day, hour of week, turn); the last column, the preferred Study 3 design, replaces day and hour of
week with exact clock hour.
The Study 2 instrument has a standard deviation of \StwoInstrSdRaw log units (\StwoInstrSdResid after the fixed
effects); the Study 3 instrument, \SthreeLoadSd. Standard errors clustered two ways (Study 2: visitor and day;
Study 3: hashed IP and clock hour).}
\end{table}

%% file: generated/tab_s2_robust.tex
\begin{table}[t]
\centering
\caption{Study 2: robustness of the follow-up result (percentage points per log unit of slowness unless stated).}
\label{tab:s2robust}
\small
\begin{tabular}{lcc}
\toprule
Dependent variable: asks a second question & OLS & IV \\
\midrule
Main specification (Table~\ref{tab:s2engage}) & $-$1.8 (0.2) & $-$2.0 (1.0) \\
Instrument leaves out only the focal conversation &  & $-$1.9 (1.0) \\
Standard errors clustered by visitor only &  & $-$2.0 (0.8) \\
Standard errors clustered by visitor and week & $-$1.8 (0.2) & $-$2.0 (1.1) \\
Within visitor, clustered by visitor and week &  & $-$2.4 (1.2) \\
Without length and reasoning controls in the regression & $-$2.2 (0.2) & $-$2.0 (1.0) \\
Regressor: log generation time, not net of length &  & $-$2.0 (1.0) \\
Reduced form: congestion instrument itself &  & $-$1.8 (0.9) \\
Regressor: log seconds of the longer wait & $-$1.9 (0.2) & $-$2.2 (1.1) \\
No reasoning model in the pair & $-$1.6 (0.3) & $-$1.5 (1.0) \\
Visitors on their first conversation & $-$2.1 (0.3) &  \\
\midrule
Slowest fifth of conversations (indicator) & $-$2.3 (0.3) & \\
Slope below the 80th percentile of slowness & 0.2 (0.3) & \\
Slope above the 80th percentile of slowness & $-$5.1 (0.4) & \\
\bottomrule
\end{tabular}
\par\smallskip
\parbox{0.95\linewidth}{\footnotesize Notes: Fixed effects, controls and two-way clustering as in
Table~\ref{tab:s2engage} unless stated. The reduced-form row reports the effect of one log unit of the congestion
instrument. In the row without length and reasoning controls, slowness and the instrument are still measured net of
length; in the row with log generation time as the regressor, the instrument is unchanged and the regression controls
for length and reasoning length. The indicator and spline rows are OLS with the same fixed effects; the 80th
percentile of slowness is
0.41 log units.}
\end{table}

%% file: generated/tab_s2_topics.tex
\begin{table}[t]
\centering
\caption{Study 2: slowness and the blind vote by topic, and qualifier tags (OLS, percentage points per log unit).}
\label{tab:s2topics}
\small
\begin{tabular}{lcc}
\toprule
 & Slower answer & Decisive votes \\
\midrule
\multicolumn{3}{l}{\emph{The slower answer wins, by topic of the conversation}} \\
Society & $-$8.7 (4.1) & 1{,}680 \\
Science \& technology & $-$3.5 (1.0) & 15{,}513 \\
Education & $-$2.1 (0.8) & 19{,}551 \\
Personal \& daily life & $-$1.7 (1.7) & 7{,}531 \\
Business \& finance & $-$0.9 (1.7) & 7{,}514 \\
Health \& environment & 0.2 (1.9) & 5{,}821 \\
Politics \& law & 0.6 (1.6) & 7{,}603 \\
Entertainment \& hobbies & 1.1 (1.5) & 8{,}303 \\
Culture \& arts & 4.1 (2.0) & 5{,}184 \\
Other & 4.3 (5.7) & 1{,}505 \\
\midrule
\multicolumn{3}{l}{\emph{Qualifier tags (left minus right), all votes}} \\
Useful & 0.3 (0.3) & \\
Complete & 1.0 (0.4) & \\
Clear formatting & $-$0.7 (0.4) & \\
Creative & 0.1 (0.2) & \\
Superficial & $-$0.1 (0.3) & \\
Instructions not followed & 0.1 (0.2) & \\
Incorrect & 0.4 (0.2) & \\
\bottomrule
\end{tabular}
\par\smallskip
\parbox{0.95\linewidth}{\footnotesize Notes: OLS with unordered model pair $\times$ month fixed effects and length and
reasoning controls (Table~\ref{tab:s2votes}), estimated separately by topic (topics with at least 1,500 decisive
votes); exploratory, without adjustment for multiple comparisons. Standard errors clustered by visitor and day.}
\end{table}